\pdfoutput=1
\documentclass[11pt]{article}
\usepackage[final]{acl}
\usepackage{times}
\usepackage{latexsym}

\usepackage[T1]{fontenc}
\usepackage[utf8]{inputenc}
\usepackage{xcolor}
\usepackage{microtype}
\usepackage{graphicx}
\usepackage{booktabs}
\usepackage{amsmath,amssymb}
\usepackage{multirow}
\usepackage{array}
\usepackage{listings}
\usepackage[hyphens]{xurl}

\newcommand{\merlin}{\textsc{MERLIN-DDx}}
\newcommand{\merlinshort}{\textsc{Merlin}}
\newcommand{\mimic}{\textsc{MIMIC-IV}}
\newcommand{\clinibench}{\textsc{CliniBench}}
\newcommand{\wikidoc}{\textsc{WikiDoc}}

\newcommand{\snd}[1]{\underline{#1}}

\newcommand{\sd}[1]{\,{\scriptsize$\pm$#1}}
\newcommand{\Vone}{$V_1$}
\newcommand{\Vtwo}{$V_2$}
\newcommand{\Vthree}{$V_3$}
\newcommand{\Vfour}{$V_4$}

\title{\merlin{}: Learning Clinical Diagnostic Reasoning\\ from Reference-Filtered Traces}

\author{
 Jan Frick\textsuperscript{1,*,\dag},
 Dennis Fast\textsuperscript{1,*},
 Conor Fallon\textsuperscript{1,*},
 Konstantin Schlüter\textsuperscript{2},
 Roseli Neuschäfer\textsuperscript{3},\\
 \textbf{Jonas Antonio Cittadino\textsuperscript{1},
 Felix Gers\textsuperscript{1},
 Alexander Löser\textsuperscript{1}}\\
 \textsuperscript{1}Berliner Hochschule für Technik,
 \textsuperscript{2}Charité -- Universitätsmedizin Berlin,\\
 \textsuperscript{3}Justus Liebig Universität Gie\ss{}en
}

\begin{document}
\maketitle
{\renewcommand{\thefootnote}{}\footnotetext{\textsuperscript{*}Equal contribution. \textsuperscript{\dag}Project lead.}}

\begin{abstract}
Large language models reason step by step in mathematics and code, but in clinical differential diagnosis (DDx) they still map notes straight to answers. One reason is that the supervision needed to teach clinical reasoning is largely missing, because clinical notes record conclusions rather than the reasoning traces behind them. We present a four-stage generator--verifier pipeline that builds this supervision. It generates diagnostic reasoning traces over real hospital admissions and checks the structured answer of every intermediate step against an external clinical reference (\wikidoc{}) and the recorded patient outcome (\mimic{}). The result is \merlin{}, a dataset of $124{,}510$ stage records over $10{,}400$ admissions{} whose discharge diagnosis \wikidoc{} covers, of which $117{,}580$ carry a verifier score; clinicians confirm the reference diagnosis in $77\%$ of reviewed cases (inter-rater $\kappa=0.91$). We evaluate two admission-time tasks, differential diagnosis and ICD coding, predicting discharge codes from what was known on arrival. This is harder than coding finished discharge summaries, and closer to real decision support. Fine-tuning Qwen3 models (0.6B--32B) on \merlin{} improves long-tail ICD performance (macro-F1) over the untuned base at every scale. Case-matched label-only controls, trained on the same cases and gold codes without traces, isolate what the traces add: $1.7$ to $6.9$ macro-F1 points at every scale, though not consistently on micro-F1. A data ablation at 8B shows macro-F1 rising with verifier-filter strictness, although stricter filtering retains fewer traces. An error analysis shows the gain reaches every frequency stratum and is especially large on low-frequency codes, which are improved but not solved. We release the dataset, pipeline, and prompts.
\end{abstract}

\newpage %
\section{Introduction}
\label{sec:intro}

Large language models (LLMs) achieve strong results in coding, mathematics, science, and law \citep{muennighoff_s1_2025,LIMO,OPENTHOUGHTS, cai-etal-2025-unilaw}, where explicit intermediate reasoning has improved performance on multi-step tasks.
In clinical differential diagnosis (DDx), LLMs still generate diagnoses directly from textual cues, without exposing the inferences that connect evidence to conditions \citep{zhou2025large, qiu2025quantifying}.
That matters in medicine, where diagnostic errors frequently involve faulty synthesis of available information \citep{graber2005diagnostic}: accuracy alone is insufficient if the output cannot be inspected and challenged.
Encoder classifiers, the incumbent for this task, expose no intermediate steps either \citep{protopatient_van-aken-2022,kayser-etal-2024-fool}, and both families remain prone to superficial shortcuts \citep{elangovan-etal-2024-principles}.

\begin{figure}[t]
\centering
\includegraphics[width=\columnwidth]{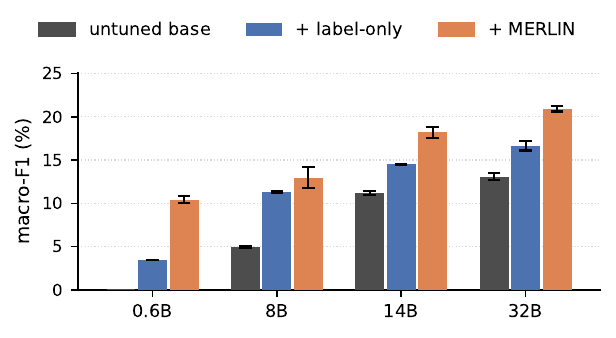}
\caption{\textbf{Fine-tuning on \merlin{} improves ICD macro-F1 at every scale, beyond a label-only control.} Untuned Qwen3 bases compared with the same models fine-tuned on \merlin{} traces or on a case-matched label-only control (same cases and codes, no traces), evaluated on the test split using our primary metric. Means over three evaluation repeats (8B fine-tuned bars: three training seeds), whiskers std. The 0.6B base bar is $0.1$.}
\label{fig:mainbars}
\end{figure}

\begin{figure*}[t]
\centering
\includegraphics[width=0.98\textwidth]{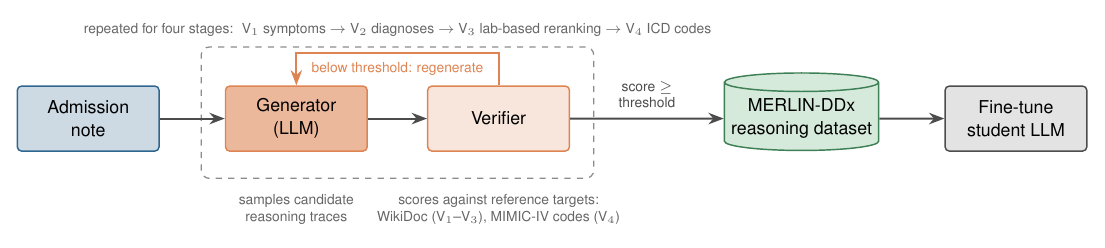}
\caption{\textbf{The \merlin{} generator--verifier pipeline.} An admission note enters a loop in which a generator LLM samples candidate reasoning traces and a verifier scores each against an external reference target. A trace scoring at or above that stage's threshold passes on; below it, the stage regenerates, until the round budget is spent and the stage falls back to a label-only target. The loop runs over four stages: symptom extraction (\Vone{}), diagnosis ranking (\Vtwo{}), lab-based reranking (\Vthree{}) and ICD-code prediction (\Vfour{}). Accepted stage outputs and label-only fallbacks are assembled into the \merlin{} dataset and used to fine-tune a student model. Our contribution is the construction loop rather than any individual stage. Because each accepted trace retains its verifier score, the released dataset can be re-filtered using alternative thresholds (\S\ref{sec:ablations}). Prompts, thresholds and per-stage verifier metrics: \S\ref{sec:pipeline_stages} and Appendix~\ref{app:prompts}.}
\label{fig:generation}
\end{figure*}

We introduce a four-stage generator--verifier pipeline that generates intermediate traces aligned with the diagnostic workflow, checks them against an external clinical reference and the recorded outcome, and runs supervised fine-tuning (SFT) on the traces that pass. We use three related terms throughout: \emph{the pipeline} is the method and the contribution (\S\ref{sec:methodology}); \merlin{} is the resulting dataset derived from $10{,}400$ \mimic{} admissions, which we release; and $+$\,\merlinshort{} marks a Qwen3 base model fine-tuned on that dataset (\S\ref{sec:results}). In this work, reasoning traces provide structured intermediate supervision aligned with the clinical diagnostic workflow.

We evaluate admission-time diagnosis ranking and ICD-code prediction. The input consists only of notes available at admission, while the labels are the discharge diagnoses recorded over the course of the stay \citep{van_aken_clinical_2021}. This is harder than coding from discharge summaries, which already contain the diagnostic narrative, and more closely reflects the point at which diagnostic support would be used. Because the ICD label space is large and highly imbalanced, we use macro-F1 over ICD categories as our primary metric: it weights each category equally rather than allowing frequent codes to dominate the score. 

\paragraph{Contributions.}
\textbf{(1)} A four-stage generator--verifier pipeline aligned with the diagnostic workflow, in which per-stage acceptance thresholds act as a \emph{compute-budget mechanism} concentrating regeneration on difficult cases. \textbf{(2)} \merlin{}, the dataset it produces: $124{,}510$ stage records over $10{,}400$ real \mimic{} admissions, $117{,}580$ scored against stage-specific reference targets, with $68\%$ of train-split records clearing their stage threshold, released with code, prompts and splits\footnote{\url{https://github.com/DATEXIS/merlin}} (dataset access follows PhysioNet credentialing upon publication). \textbf{(3)} An empirical study that separates two effects. Fine-tuning on \merlin{} improves ICD macro-F1 over untuned bases at every scale (0.6B--32B). Case-matched label-only controls at every scale isolate the traces' contribution, which holds for macro-F1 but not consistently for micro-F1. \textbf{(4)} An analysis showing that macro-F1 rises with verifier-filter strictness rather than with trace volume, reflects recovery of documented diagnoses and comorbidities rather than over-prediction, survives removing patients shared between splits, and is proportionally larger for low-frequency codes; a clinician study additionally assesses the clinical plausibility of the reference targets. A clinician audit finds that the verifier loop does not improve rationale quality.

\section{Background \& Related Work}
\label{sec:related}

\paragraph{LLM reasoning in formal vs.\ clinical domains.}
Reasoning frameworks such as S1 \citep{muennighoff_s1_2025}, LIMO \citep{LIMO}, and DeepSeek-R1 \citep{deepseek-ai_deepseek-r1_2025} show that verification signals can stabilize multi-step generation. \citet{kambhampati_stop_2026} argue that intermediate tokens should not be read as thought. In medicine verification must reflect clinical plausibility rather than formal correctness, yet the reasoning linking symptoms to diagnoses is rarely documented.

\paragraph{Clinical reasoning LLMs.}
HuatuoGPT-o1 \citep{chen_huatuogpt-o1_2024}, FineMedLM-o1 \citep{yu_finemedlm-o1_2025}, and MedReason \citep{wu_medreason_2025} extend general reasoning to medical tasks but retain intermediate steps without comparing them against an external resource before training. Others inject reasoning via rationale-enhanced representations \citep{liu_carer_2024}, knowledge-guided retrieval \citep{song_kare_2025}, or agentic EHR reasoning \citep{xiong_ehragent_2024}, and closed systems such as AMIE \citep{tu_towards_2024} do not release the model and training assets required for exact reproduction. Process Reward Agents \citep{sohn_pra_2026} share our premise that clinical intermediate steps are not verifiable without external knowledge, but apply it online, steering a frozen policy at decode time. We apply it offline, to build a reusable dataset. The closest prior resource, \mimic{}-Ext-DiReCT \citep{wang_direct_2024}, provides $511$ physician-annotated admission-note chains checked against a diagnostic knowledge graph, but is evaluation-only and covers $5\%$ as many admissions as ours. Closest in setting is DeepICD-R1 \citep{rhr-etal-2026-deepicd}, which also distills and releases reasoning traces over \mimic{} admission notes. It predicts only the first-listed diagnosis code and rewards the final code, whereas we predict all discharge codes and check every intermediate step against an external reference. Contemporary work post-trains LLMs for generative ICD coding from discharge summaries \citep{northwestern_coders_2026} or trains evidence-seeking diagnosis with RL without releasing a dataset \citep{rl_evidence_2026}; MedKGI \citep{medkgi_2025} targets KG-grounded multi-turn diagnostic inquiry.

\paragraph{Knowledge bases and clinical verification.}
We build on \mimic{} \citep{johnson2023mimiciv} and use \wikidoc{} \citep{wikidoc}, a peer-edited clinical online encyclopedia whose structured symptom--disease tables have been used in prior medical NLP resources \citep{han2023medalpaca}. 
Unlike approaches that use external resources only for retrieval or reward \citep{song_kare_2025}, we use them to check intermediate steps. \wikidoc{} is one possible reference; we discuss its limitations below.

Among benchmarks, \clinibench{} \citep{grundmann2025clinibench} evaluates discharge-diagnosis prediction from admission notes in \mimic{} and is what we align our task and encoder baselines with; DR.BENCH \citep{gao_drbench_2023} and MedXpertQA \citep{zuo_medxpertqa_2025} target reasoning via QA, and MSDiagnosis \citep{zhang_msdiagnosis_2024}, ER-Reason \citep{liu_erreason_2025}, and Clinical Reasoning Graphs \citep{clinreasoninggraphs_2026} structure diagnosis across stages. In contrast, \merlin{} uses verifier-checked intermediate traces as training supervision and evaluates their effect on both diagnosis ranking and ICD-code prediction.

\section{The Generator--Verifier Pipeline}
\label{sec:methodology}

The pipeline broadly mirrors the clinical diagnostic process: identifying symptoms, forming a differential diagnosis, refining it using additional clinical evidence such as laboratory results, and assigning discharge ICD codes.
Sections~\ref{sec:clinical_grounding}--\ref{sec:dynamics} describe the \emph{pipeline} itself, an instruction--generation--verification loop (Figure~\ref{fig:generation}) that validates each response against an external reference target; \S\ref{sec:dataset} describes the \emph{dataset} it produced and that we release.

\subsection{What the Verifiers Check Against}
\label{sec:clinical_grounding}
To check intermediate steps we map each \mimic{} case to the \wikidoc{} differential-diagnosis table for its chief complaint, and match its primary discharge diagnosis to the corresponding disease row. That row supplies two targets: the disease-specific list of signs and symptoms, used for symptom extraction, and the associated differential-diagnosis list, used for diagnosis ranking. Every verifier except ICD coding, which scores against the \mimic{} discharge codes, derives from this mapping, so we assess the clinical plausibility of the retained targets with a clinician study (\S\ref{sec:annotation}). Automatic mapping scores and the filtering used to construct the retained sample are reported by chief complaint in Appendix Figure~\ref{fig:match}.

\subsection{The Four Pipeline Stages}
\label{sec:pipeline_stages}
Each stage pairs a generation prompt with a verifier metric (full prompts in Appendix~\ref{app:prompts}; context assembly and a worked example in Appendix Figure~\ref{fig:pipeline_impl}). Every verifier scores against the confirmed diagnosis or the discharge codes: label information builds the dataset and is never available at evaluation time (\S\ref{sec:experimental_setup}).

\paragraph{Symptom extraction.} The model extracts a structured symptom profile from the admission note. We represent predictions and \wikidoc{} references as vectors (present $+1$, absent $-1$, not mentioned $0$) and score candidates with a \mbox{normalized} dot product,
\begin{equation*}
\mathrm{NormDot}(p,\ell)=\frac{p^\top \ell + \|\ell\|_0}{2\|\ell\|_0},
\end{equation*}
which scales from $0$ (contradiction) to $1$ (perfect agreement); traces are accepted at $\mathrm{NormDot}\ge 0.60$. \wikidoc{}'s variable ($\pm$) entries are encoded as $0$, and an empty extraction scores $0.5$. The reference $\ell$ is the symptom list \wikidoc{} associates with the confirmed disease, so NormDot measures agreement with a disease-level prototype rather than per-note extraction accuracy. We therefore use it as a construction-time gate, not an evaluation metric, and report its test-split values in Appendix~\ref{app:symptom}. Symptoms outside the prototype leave the score unchanged, so an extraction can keep findings that point to other diagnoses.

\paragraph{Diagnosis ranking.} The model forms an initial differential from the extracted symptoms, cross-checked against the mapped \wikidoc{} differential. Candidates are scored by the reciprocal rank (RR) of the confirmed diagnosis and accepted at $\mathrm{RR}\ge 0.50$. Passing establishes that the confirmed diagnosis is ranked first or second. 

\paragraph{Lab-based reranking.} Laboratory results materially refine a differential \citep{bhasuran2025preliminary}. The resulting training traces provide supervised examples of how laboratory evidence revises a differential. We include results drawn within $12$ hours of admission \citep{cifra2020factors}: wide enough to cover tests ordered on arrival and the first morning draw, narrow enough to exclude most investigations ordered in response to a later working diagnosis. Acceptance is one rank tighter: reranked differentials are accepted at $\mathrm{RR}\ge 0.80$, so the confirmed diagnosis leads the differential once labs are seen. Lab-based reranking is used during dataset construction only (\S\ref{sec:experimental_setup}).

\paragraph{ICD-code prediction.} The final stage maps all preceding evidence to ICD codes, truncated to three characters as in prior admission-time work on this \citep{van_aken_clinical_2021,grundmann2025clinibench}, scored against the \mimic{} discharge codes with micro-F1 and accepted at $\mathrm{F1}_{\mathrm{micro}}\ge 0.55$. Passing means the predicted three-character categories largely match the discharge codes.

\paragraph{Where the thresholds come from.}
Each threshold sits at or just above its stage's mean verifier score, a fixed rule that spends the regeneration budget where the generator struggles rather than a value tuned on downstream performance. A score above threshold means a trace agrees with its reference target, not that it is clinically correct. The ICD cutoff sits furthest above its stage mean, because it gates the final, evaluated output. We evaluate this choice in \S\ref{sec:ablations} by comparing it with a looser half-threshold variant. 
\subsection{Generator and Verifier Dynamics}
\label{sec:dynamics}
Given a \mimic{} admission note and stage-specific instructions, the generator samples $30$ candidate traces per round as structured JSON and keeps the best-scoring one; if it falls below the stage threshold the round repeats, up to $4$ rounds, and candidates that never pass are replaced with a no-reasoning fallback (\S\ref{sec:experimental_setup}). Evaluation is much cheaper: $4$ candidates per stage, first valid JSON retained. To reduce dependence on any single generator, we use three models: Qwen3-32B~\citep{yang_qwen3_2025}, MedGemma-27B~\citep{sellergren_medgemma_2025}, and Llama-3.3-70B-Instruct~\citep{grattafiori_llama3_2024}. In the first round, $51.6$, $59.5$, $62.2$ and $22.3\%$ of admission runs pass symptom extraction, diagnosis ranking, lab reranking and ICD coding; after regeneration, $64.1$, $82.8$, $68.6$ and $37.2\%$ enter the dataset with a trace and the rest as label-only fallbacks (over all generation runs; train-split rates in Appendix Table~\ref{tab:counts}). The loop costs $10.0$ generation prompts per admission instead of $4.0$, and $63.9\%$ of generation wall-clock is spent in retry rounds. The loop also improves structural validity, from $21.1\%$ initial JSON validity for MedGemma-27B to $80$--$97\%$ across all three by the final dataset.

\subsection{The Resulting Dataset}
\label{sec:dataset}
We construct simulated admission notes from ICD-10-coded admissions in the \mimic{} v2.2 \texttt{hosp} subset and their discharge summaries in \mimic{}-Note v2.2 following \citet{van_aken_clinical_2021}, sampling across the seven most frequent chief complaints by admission count. Filtered chains become instruction--response pairs for SFT; before training, accepted answers scoring below $1$ are reconciled with the reference. At the coding stage this changes over $99\%$ of accepted traces and makes $34$--$40\%$ of their per-code reasons placeholders, so every coding target lists the gold three-character categories, as in the label-only data, and differs from it in the reasoning and the per-code reasons. Each reasoning block closes with a templated summary of the model's own, unreconciled answer (Appendix~\ref{app:prompts}). The final dataset covers $10{,}400$ admissions (train $7{,}299$; dev $917$;
test $2{,}184$), each run through the pipeline by all three generators, giving
$31{,}200$ four-stage chains and $124{,}510$ stage-level records. Of these,
$117{,}580$ carry a verifier score; the remaining $5.6\%$ are generations whose
output did not validate against the stage schema and therefore cannot be scored.
These are overwhelmingly stage V1 ($88\%$ of unscored records), where the schema
requires the complete complaint-specific manifestation checklist and a partially
filled checklist is rejected, and are concentrated in a single generator
(MedGemma-27B fails V1 on $56.8\%$ of admissions, versus $4.9\%$ for
Llama-3.3-70B and $0\%$ for Qwen3-32B; Appendix~\ref{app:counts}). Cases are multi-label, with a median of $12$ discharge ICD codes, and the label space spans $1{,}157$ distinct three-character codes with a pronounced long tail (Appendix~\ref{app:longtail}). The split is admission-wise rather than patient-wise, which we analyse in \S\ref{sec:error_shift}.

\section{Experimental Setup}
\label{sec:experimental_setup}

At test time we evaluate the two stages that admission-time evidence supports and that measure diagnostic output: diagnosis ranking (\Vtwo{}) and ICD-code prediction (\Vfour{}). Lab reranking (\Vthree{}) is excluded because labs are not yet available at $t{=}0$; symptom extraction (\Vone{}) still runs, since \Vtwo{} conditions on it, but its NormDot score is reported only as a pipeline diagnostic (\S\ref{sec:pipeline_stages}). For every fine-tuned model we sweep training epochs on \emph{dev}, select the best epoch by ICD macro-F1, and report that checkpoint on the held-out \emph{test} split.

\paragraph{Baselines.}
We compare against \textbf{MedGemma-27B}, \textbf{Llama-3.3-70B-Instruct} (testing whether scale alone suffices), \textbf{Baichuan-M2-32B}~\citep{baichuan_m2_2025} (an open medical reasoning model at the scale of Qwen3-32B), the \textbf{Qwen3} base family (0.6B--32B), and \textbf{encoder classifiers} retrained on our \mimic{} subset, the well-established approach here and the only discriminative family; the strongest appears in Table~\ref{tab:main}, all three in Appendix~\ref{app:encoders}. Every row uses the identical admission-time protocol.

\paragraph{Supervised fine-tuning.}
We fine-tune Qwen3 base models on \merlin{} (ICD-coding rows duplicated, ablated in \S\ref{sec:ablations}; AdamW, 4 epochs, max length 8{,}192, seed 42; Appendix~\ref{app:training}), with LoRA reported as an adaptation-method ablation in Appendix~\ref{app:lora}. To isolate what the traces add we additionally fine-tune every base model on a label-only dataset with the same cases, instructions, ICD-row duplication and gold outputs but \emph{no} traces; this is also the fallback replacing traces that fail a verifier threshold, so all variants see the same cases.

Metrics are Recall@$k$ for diagnosis ranking and micro/macro-F1 over three-character codes for ICD prediction, scored by the same verifiers used in generation; macro-F1 is primary (\S\ref{sec:intro}). Macro-F1 averages per-code F1 over the $830$ three-character codes in the test gold labels; predicted codes outside this set are discarded before computing any ICD metric. Counting them as false positives instead lowers macro-F1 by up to $3.1$ points and reverses no \merlinshort{}--label-only comparison (Appendix~\ref{app:longtail}).

\section{Results}
\label{sec:results}

\begin{table*}[t]
\centering
\small
\setlength{\tabcolsep}{6pt}
\renewcommand{\arraystretch}{1.1}
\begin{tabular}{l cc @{\hskip 18pt} cccc}
\toprule
 & \multicolumn{2}{c}{\textbf{Diagnosis ranking}} & \multicolumn{4}{c}{\textbf{ICD codes}} \\
\cmidrule(lr){2-3} \cmidrule(lr){4-7}
\textbf{Model} & \textbf{R@1} & \textbf{R@3} & $\mathbf{F1}_{\mathrm{mic}}$ & $\mathbf{F1}_{\mathrm{mac}}$ & $\mathbf{R}_{\mathrm{mac}}$ & $\mathbf{P}_{\mathrm{mac}}$ \\
\midrule
\multicolumn{7}{@{}l}{\emph{Untuned baselines}}\\
MedGemma-27B & \underline{$51.8$}\sd{0.2} & $74.5$\sd{0.3} & $29.4$\sd{0.2} & $7.6$\sd{0.1} & $9.1$\sd{0.3} & $11.8$\sd{0.3} \\
Llama-3.3-70B-Instruct & $51.1$\sd{0.5} & $72.5$\sd{0.6} & $29.7$\sd{0.6} & $11.8$\sd{0.7} & $14.4$\sd{0.5} & $18.0$\sd{0.4} \\
Baichuan-M2-32B & $49.6$\sd{0.3} & $71.8$\sd{0.7} & $32.4$\sd{1.1} & $11.1$\sd{0.9} & $14.8$\sd{1.1} & $13.2$\sd{1.0} \\
\addlinespace[4pt]
\multicolumn{7}{@{}l}{\emph{Qwen3 base and fine-tuned}}\\
Qwen3-0.6B & $34.2$\sd{0.3} & $52.9$\sd{0.4} & $1.2$\sd{0.1} & $0.1$\sd{0.0} & $0.1$\sd{0.0} & $0.4$\sd{0.0} \\
\quad$+$\,label-only$^{\dagger}$ & $11.4$\sd{2.9} & $13.5$\sd{3.3} & $33.4$\sd{0.3} & $3.5$\sd{0.0} & $3.6$\sd{0.1} & $4.7$\sd{0.4} \\
\quad$+$\,\merlinshort{} & $51.0$\sd{0.5} & $73.1$\sd{0.7} & $43.6$\sd{0.7} & $10.4$\sd{0.4} & $10.4$\sd{0.4} & $12.5$\sd{0.6} \\
\addlinespace[2pt]
Qwen3-8B & $45.5$\sd{0.5} & $66.6$\sd{0.5} & $19.7$\sd{0.1} & $5.0$\sd{0.1} & $4.9$\sd{0.2} & $9.1$\sd{0.2} \\
\quad$+$\,label-only$^{\dagger}$$^{\ast}$ & $51.2$\sd{1.4} & $68.7$\sd{1.3} & $44.1$\sd{0.6} & $11.3$\sd{0.1} & $10.9$\sd{0.1} & $14.6$\sd{0.1} \\
\quad$+$\,\merlinshort{}$^{\ast}$ & $\mathbf{53.9}$\sd{0.5} & \underline{$76.3$}\sd{0.4} & $39.8$\sd{4.2} & $13.0$\sd{1.3} & $13.3$\sd{0.9} & $17.8$\sd{0.9} \\
\addlinespace[2pt]
Qwen3-14B & $49.0$\sd{0.1} & $68.9$\sd{0.3} & $30.7$\sd{0.3} & $11.2$\sd{0.2} & $14.0$\sd{0.3} & $14.6$\sd{0.3} \\
\quad$+$\,label-only$^{\dagger}$ & $50.5$\sd{2.2} & $65.3$\sd{2.4} & $47.0$\sd{0.3} & $14.5$\sd{0.1} & $14.9$\sd{0.2} & $17.6$\sd{0.4} \\
\quad$+$\,\merlinshort{} & $46.6$\sd{1.3} & $68.7$\sd{4.7} & \underline{$48.8$}\sd{0.5} & \underline{$18.2$}\sd{0.6} & \underline{$17.9$}\sd{0.8} & \underline{$21.5$}\sd{0.7} \\
\addlinespace[2pt]
Qwen3-32B & $50.5$\sd{0.7} & $70.3$\sd{0.2} & $33.6$\sd{0.1} & $13.1$\sd{0.4} & $16.2$\sd{0.6} & $15.9$\sd{0.6} \\
\quad$+$\,label-only$^{\dagger}$ & $27.9$\sd{5.6} & $40.4$\sd{8.2} & $47.3$\sd{0.5} & $16.6$\sd{0.5} & $17.1$\sd{0.3} & $19.7$\sd{0.8} \\
\quad$+$\,\merlinshort{} & $36.0$\sd{1.5} & $\mathbf{76.8}$\sd{1.5} & $\mathbf{50.7}$\sd{0.2} & $\mathbf{20.9}$\sd{0.3} & $\mathbf{20.9}$\sd{0.5} & $\mathbf{23.6}$\sd{0.1} \\
\addlinespace[4pt]
\multicolumn{7}{@{}l}{\emph{Encoder classifier}}\\
BioClin.\ ModernBERT$^{\ddagger}$ & -- & -- & $26.4$\sd{0.8} & $8.2$\sd{0.1} & $13.8$\sd{2.3} & $7.0$\sd{0.7} \\
\bottomrule
\end{tabular}
\caption{\textbf{Main results on the held-out test split} ($n{=}2{,}184$), all rows under the same admission-time protocol: no lab values, no reranking, dev-selected best epoch. \textbf{Bold} marks the best value in each column and \snd{underline} the second best; $\mathrm{F1}_{\mathrm{mac}}$ is the primary metric (\S\ref{sec:intro}). Rows marked $^{\ast}$ are mean\,$\pm$\,std over three training seeds, each evaluated once; every other model row is mean\,$\pm$\,std over three evaluation repeats of one checkpoint, capturing decoding stochasticity but \emph{not} training-seed variance (\S\ref{sec:robustness}). \textbf{R@1}/\textbf{R@3} are the shares of cases whose confirmed diagnosis is ranked first / in the top three (the two come apart at 32B, \S\ref{sec:ranking}). ICD columns are micro- and macro-F1 over three-character codes, then macro-averaged recall and precision. $^{\dagger}$Label-only control: same cases, instructions, gold outputs and ICD-row oversampling, traces removed. At 0.6B and 32B the label-only controls produce valid diagnosis-stage JSON for only about $21\%$ and $46\%$ of cases, which lowers their R@1 and R@3. $^{\ddagger}$Strongest of three encoder classifiers retrained on our \mimic{} subset (\ref{app:encoders}); its $\pm$ is over three \emph{training} seeds and so is the stricter quantity, and it emits no ranked differential, hence the dashes. LoRA is in \ref{app:lora}.}
\label{tab:main}
\end{table*}

\subsection{Fine-Tuning Gains and the Trace Effect}
Table~\ref{tab:main} summarizes test-split performance and Figure~\ref{fig:mainbars} plots it. On the primary metric, ICD macro-F1, fine-tuning on \merlin{} traces lifts the size-matched base at every scale, and micro-F1 moves with it. The absolute macro-F1 gain stays at $7$--$9$ points from 8B to 32B, so the improvement persists across model sizes. This comparison includes adaptation to the task and its output format; the case-matched label-only controls separate what the traces add.

\paragraph{The fine-tuned 8B model exceeds larger untuned baselines.}
Qwen3-8B fine-tuned on \merlin{} exceeds every untuned baseline we evaluated, including Llama-3.3-70B-Instruct and the open medical reasoning model Baichuan-M2-32B, despite using $3$--$9\times$ fewer parameters. Two of our three trace generators are themselves untuned baselines here, so this comparison also captures distillation from the trace generators. At 32B, the comparison is less confounded by differences between models: Qwen3-32B is itself a generator (\S\ref{sec:dynamics}), so its two rows measure one model before and after training on verifier-filtered traces, some of them its own. The 0.6B result illustrates the limitation of micro-F1. Its micro-F1 beats all three untuned baselines and exceeds the 8B \merlinshort{} model's own, while its macro-F1 lags. 

\paragraph{Comparison with encoder baselines.}
The strongest encoder reaches $8.2$ macro-F1, ahead of untuned MedGemma-27B and of the
untuned Qwen3-8B base \merlinshort{} is trained from. A $150$M supervised classifier
therefore outperforms a $27$B zero-shot generative model. Only the 14B and 32B bases
clear the encoder untuned ($11.2$ and $13.1$). Fine-tuning on \merlin{} clears it at every size. Our encoders are retrained on the
same $10{,}400$-admission subset, so this is a comparison at \emph{matched data}.
The full-data encoder remains stronger on macro-F1: scoring \clinibench{}'s trained
encoders \citep{grundmann2025clinibench} on our own test admissions, a GatorTronS-345M
trained on $50$--$80\times$ more admissions reaches $25.5$ macro-F1 against
\merlinshort{}-32B's $20.9$ (Appendix~\ref{app:encoders}). It predicts $33.6$ codes per
admission, three times as many as \merlinshort{}-32B, and reaches lower micro-F1
($29.3$ vs.\ $50.7$).

\paragraph{Effect of reasoning traces.}
\merlinshort{} outperforms the matched label-only controls on ICD macro-F1 at every scale (Table~\ref{tab:main}). At 8B, where both arms were retrained under three training seeds, the weakest \merlinshort{} seed still exceeds the strongest label-only seed (\S\ref{sec:robustness}), and a paired case-level bootstrap on the seed-42 checkpoints ($2{,}000$ resamples, so case sampling only) puts the gap at $\Delta{=}{+}2.5$ ($13.9$ vs.\ $11.4$), 95\% CI $[{+}1.6,{+}2.9]$. On micro-F1, the traces help at 0.6B, 14B and 32B, but label-only leads at 8B across seeds. \merlinshort{} also keeps the confirmed diagnosis in its top three more often than label-only at every scale.
\begin{figure*}[!t]
\centering
\includegraphics[width=0.88\textwidth]{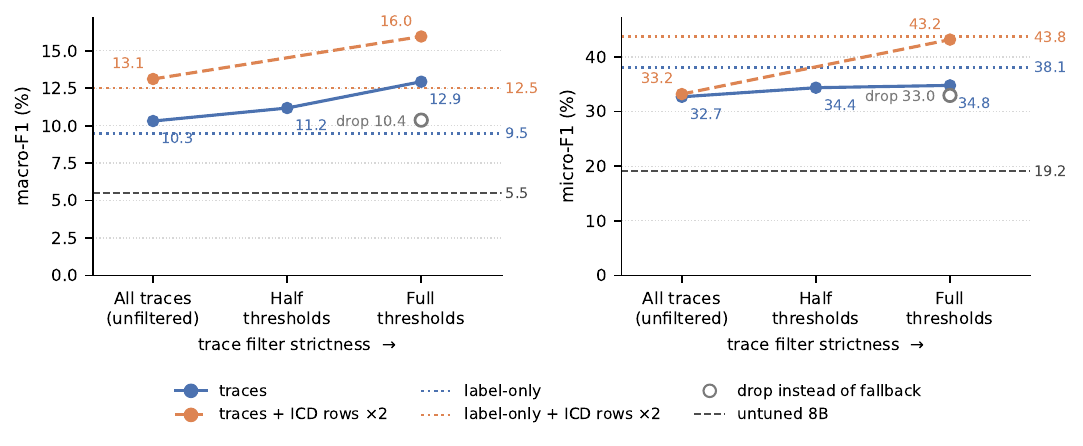}
\caption{\textbf{Filtering traces by verifier score and oversampling ICD rows each improve long-tail ICD performance.} Qwen3-8B, full fine-tuning, dev split, dev-selected best epoch. Left to right: label-only (no traces), traces with no score threshold, then traces clearing half and all stage thresholds. Blue is without and orange with duplicated ICD-coding rows. The open circle drops below-threshold cases instead of using the label-only fallback; half thresholds were not run with ICD-row duplication. Macro-F1 (left) rises with filter strictness at both oversampling settings, while micro-F1 (right) is matched by oversampling alone. Diagnosis Recall@3 stays within two points of the best result across the trace-based variants. Exact values, including the two controls the markers do not separate, in Appendix~\ref{app:ablation}.}
\label{fig:ablation}
\end{figure*}

\subsection{Diagnosis Ranking}
\label{sec:ranking}
Diagnosis ranking improves clearly at the two smaller scales and is flat at 14B (Table~\ref{tab:main}).

\paragraph{Recall@1 and Recall@3 diverge at 32B.}
 
Fine-tuning improves both at the smaller scales, and 8B gives the best rank-1 score in Table~\ref{tab:main}, $2.7$ points above label-only. At 32B they diverge: Recall@1 falls by $14.5$ points while Recall@3 rises to the best in the table, both moves far outside eval-repeat variance. Fine-tuned differentials converge to roughly the same length at every scale while the 32B base starts nearly twice as long as the smaller bases, which is consistent with the confirmed diagnosis staying in the top three while slipping out of first place there. The divergence is a trade-off between identifying the leading diagnosis and keeping it in a shortlist.

\paragraph{Selecting on ICD macro-F1 has a cost.}
Over the dev sweep, ICD macro-F1 keeps rising into late epochs while diagnosis ranking and structured output degrade (Figure~\ref{fig:epoch_tradeoff}): mild at 0.6B and 8B, severe at 14B, where macro-F1 climbs by nearly six points between epochs 1 and 3 while diagnosis-stage JSON validity falls from $100\%$ to $82.9\%$ and Recall@3 falls with it. The two track each other mechanically: an unparseable diagnosis stage scores zero on Recall@$k$. Conditioning on parseable output holds 14B Recall@3 within three points across all four epochs.

\subsection{Robustness}
\label{sec:robustness}
Outside the 8B fine-tuned rows and the encoder row, the $\pm$ values in Table~\ref{tab:main} are evaluation repeats of a single
checkpoint. To isolate \emph{training} variance we retrained the primary 8B configuration and the label-only control from scratch under two further seeds; Table~\ref{tab:main} reports their means.
Diagnosis ranking is highly stable (Recall@3 $76.3{\pm}0.4$); ICD F1 varies
more (macro $13.0{\pm}1.3$, micro $39.8{\pm}4.2$), but both spreads remain well below the base-versus-fine-tuned gaps. Across seeds the reasoning recipe
retains its macro-F1 advantage over label-only but not its micro-F1 advantage (per-seed values in Appendix Figure~\ref{fig:robustness}).

\begin{table*}[t]
\centering
\small
\setlength{\tabcolsep}{3.2pt}
\renewcommand{\arraystretch}{1.1}
\begin{tabular}{llccccw{c}{2.8em}w{c}{2.8em}w{c}{2.8em}c}
\toprule
 & & \multicolumn{4}{c}{\textbf{Recall / F1 by code class (\%)}} & \multicolumn{3}{c}{\textbf{F1 by train.\ frequency}} & \textbf{Full code} \\
\cmidrule(lr){3-6} \cmidrule(lr){7-9} \cmidrule(lr){10-10}
\textbf{Size} & & \textbf{Primary} & \textbf{Hist.\ F1} & \textbf{Med.\ F1} & \textbf{Chron.\ F1} & \textbf{Head} & \textbf{Body} & \textbf{Tail} & $\mathbf{F1}_{\mathrm{mac}}$ \\
\midrule
\multirow{3}{*}{0.6B} & base & $1.1$ & $0.0$ & $0.0$ & $3.2$ & $0.3$ & $0.1$ & $0.0$ & $0.0$ \\
 & $+$\,label-only$^{\dagger}$ & $52.1$ & $35.3$ & $56.6$ & $43.4$ & $16.3$ & $1.1$ & $0.0$ & $1.0$ \\
 & $+$\,\merlinshort{} & $63.0$ & $44.1$ & $65.1$ & $57.5$ & $31.6$ & $9.4$ & $0.6$ & $1.5$ \\
\addlinespace[2pt]
\multirow{3}{*}{8B} & base & $39.9$ & $8.5$ & $0.0$ & $35.8$ & $12.4$ & $5.5$ & $0.9$ & $0.4$ \\
 & $+$\,label-only$^{\dagger}$ & $62.1$ & $45.1$ & $65.0$ & $57.4$ & $30.8$ & $11.4$ & $2.4$ & $3.1$ \\
 & $+$\,\merlinshort{} & $64.8$ & $44.7$ & $62.1$ & $56.4$ & $32.4$ & $15.3$ & $3.6$ & $1.7$ \\
\addlinespace[2pt]
\multirow{3}{*}{14B} & base & $63.5$ & $24.5$ & $27.7$ & $47.5$ & $24.3$ & $12.0$ & $3.6$ & $1.0$ \\
 & $+$\,label-only$^{\dagger}$ & $67.1$ & $46.6$ & $68.2$ & $62.3$ & $35.7$ & $15.0$ & $4.0$ & \underline{$4.2$} \\
 & $+$\,\merlinshort{} & $65.4$ & $48.5$ & \underline{$68.6$} & \underline{$62.6$} & \underline{$37.5$} & \underline{$21.4$} & \underline{$7.5$} & $3.1$ \\
\addlinespace[2pt]
\multirow{3}{*}{32B} & base & $56.5$ & $22.4$ & $8.5$ & $51.1$ & $26.3$ & $13.2$ & $5.7$ & $1.4$ \\
 & $+$\,label-only$^{\dagger}$ & $\mathbf{70.9}$ & \underline{$52.0$} & $66.8$ & $62.2$ & $37.0$ & $17.9$ & $7.3$ & $\mathbf{5.4}$ \\
 & $+$\,\merlinshort{} & \underline{$67.5$} & $\mathbf{53.1}$ & $\mathbf{71.6}$ & $\mathbf{63.9}$ & $\mathbf{40.1}$ & $\mathbf{23.3}$ & $\mathbf{9.6}$ & $3.5$ \\
\bottomrule
\end{tabular}

\caption{\textbf{Where the ICD gain comes from}, three cuts of the base, label-only and \merlinshort{} comparison (test split, $n{=}2{,}184$, seed-42 evaluation runs). All figures are exact set operations over gold and predicted codes; definitions in Appendix~\ref{app:qa}. \textbf{Primary}: share of cases whose predicted codes cover the first gold code, taken as principal; \textbf{Hist./Med./Chron.}: class-conditional F1 over personal-history, long-term-medication and Elixhauser comorbidity codes. \emph{Frequency strata:} head ($\ge100$), body ($10$--$99$), tail ($<10$) training occurrences, scored over each stratum's $153$/$349$/$328$ test-gold codes (\S\ref{sec:longtail}). \emph{Full code:} untruncated granularity, macro-F1 only. $^{\dagger}$Label-only control of Table~\ref{tab:main}: same cases and gold codes, no traces. \textbf{Bold}: best per column, \underline{underline}: second best.}
\label{tab:qa}
\end{table*}

\section{Analysis}
\label{sec:analysis}

\subsection{When Does Reasoning Supervision Help?}
\label{sec:ablations}

We train each of the eight dataset variants of Figure~\ref{fig:ablation} once at 8B and evaluate on the \emph{dev} split.

\paragraph{Effect of filter strictness.}
Macro-F1 increases monotonically with filter strictness at both oversampling settings ($9.5 \to 10.3 \to 11.2 \to 12.9$ without ICD oversampling, $12.5 \to 13.1 \to 16.0$ with it; Table~\ref{tab:ablation}). Despite replacing many traces with the label-only fallback, stricter filtering improves macro-F1. Stronger reference alignment thus improves macro-F1 in the tested configurations, even though it retains fewer traces. Micro-F1 behaves differently. Label-only data with oversampling matches the best trace-based recipe while trailing it by $3.5$ macro points. Dropping the cases whose traces never clear their thresholds, rather than falling back to label-only, costs a further $2.5$ macro points.
This suggests that below-threshold traces should be replaced rather than the corresponding cases removed from training.

\subsection{Where the Gain Comes From}
\label{sec:error_shift}

To see \emph{how} behaviour changed we decompose errors, comparing gold and predicted three-character categories with deterministic set operations for every test case (Table~\ref{tab:qa}). \emph{Primary} asks whether predicted codes cover the first gold code. At 8B it rises from $39.9$ to $64.8$, and every code-class score rises with it, long-term medication from zero. The same pattern holds at 32B. The 14B base is the exception, already recovering the principal diagnosis unusually well, above 32B's base, so fine-tuning moves it only marginally, while the history, medication and comorbidity scores rise sharply. These are F1 scores, so a model that merely over-predicted a class would lose precision. Overall precision rises at every model size, so the higher class-level F1 cannot be explained solely by predicting more codes. From 8B, the label-only control is on par with \merlinshort{} on these classes, ahead on Primary at 14B and 32B and behind by at most $4.8$ points elsewhere, so these gains mostly reflect adaptation to the task rather than the traces.

\paragraph{Robustness to patient overlap.}
The admission-wise split (\S\ref{sec:dataset}) leaves $390$ of $2{,}184$ test admissions ($18\%$) sharing a patient with a training admission. Restricting the analysis to the $1{,}794$ non-overlapping admissions preserves the 8B improvements on history, medication and chronic-comorbidity codes. Within overlapping admissions, chronic-comorbidity codes seen in that patient's training admission are recovered more often than unseen ones ($66.8$ vs.\ $24.7$ recall), which suggests some patient-specific memorization.

\subsection{Head, Body and Tail}
\label{sec:longtail}

Macro-F1 summarizes the long tail rather than showing it. Table~\ref{tab:qa} splits the same comparison by training frequency, reporting mean per-code F1 within each stratum. The split is severe: $153$ head codes account for $82.8\%$ of all training code occurrences, while tail codes collectively account for only $2.1\%$.

Fine-tuning on \merlin{} improves all three frequency strata, with larger relative gains for body and tail codes than for head codes. At 8B, head F1 rises $2.6\times$, body $2.8\times$ and tail more than fourfold ($0.9\to3.6$); at 32B, $1.5\times$, $1.8\times$ and $1.7\times$. Against the label-only control, the traces add little on head codes ($1.6$--$3.1$ points from 8B) but more on body ($3.9$--$6.4$) and tail codes ($1.2$--$3.5$), which hold $677$ of the $830$ scored codes and so carry most of the macro-F1 gain.
  Model size has a much larger proportional effect on tail than head performance: across \merlinshort{} models, head F1 increases from $31.6$ to $40.1$ between 0.6B and 32B, whereas tail F1 increases from $0.6$ to $9.6$.

The best model nevertheless remains below $10$ F1 on codes observed fewer than ten times in training. Fine-tuning and larger models both improve these codes, but neither compensates for their limited training frequency. 

\subsection{Clinician Studies of Targets and Rationales}
\label{sec:annotation}
Because the V1--V3 verifier signals derive from the \wikidoc{} mapping, three clinician co-authors independently reviewed cases from a $350$-case stratified frame and confirmed the clinically correct diagnosis ($\kappa=0.913$ on $236$ cases annotated by the first two reviewers, Appendix~\ref{app:annotation}). The diagnosis confirmed by the first reviewer, who covered all $350$ cases, agrees with our ICD-derived reference in $77.1\%$ of cases and with the cosine-similarity reference in $68.0\%$. These rates assess the plausibility of the reference targets, not a clinician--model comparison. Separately, the fixed \wikidoc{} candidate list omits the confirmed diagnosis in $11.4\%$ of cases, capping candidate-list recall at $0.886$. A second, blinded audit rated generated rationales from $42$ admissions (Qwen3-32B, Appendix~\ref{app:audit}). Diagnosis-stage rationales were judged clinically valid in about $92\%$ of cases and grounded in the note in at least $95\%$; at the coding stage, $67\%$ supported the specific ICD code. These levels were the same with and without the verifier loop (see Limitations).

\section{Conclusion and Future Work}
\label{sec:conclusion}

We introduced \merlin{}, $124{,}510$ stage records from $10{,}400$ admissions built by a four-stage generator--verifier pipeline. Fine-tuning on it improves ICD category macro-F1 at every tested scale, with consistent gains over matched label-only controls, and the 8B ablation identifies reference filtering and ICD-row oversampling as complementary construction choices. Better prediction did not come with measurably better rationales. Future work should identify which properties of the supervision drive learning and how to improve prediction and reasoning together.

\section*{Limitations}
\label{sec:limitations}
\emph{Scope of verification.} We check that intermediate steps agree with external reference targets, not that the reasoning is logically valid or reconstructs clinician cognition \citep{kambhampati_stop_2026}. Traces are synthetic, admission-time notes are simulated from \mimic{} following prior work, and ICD codes are truncated to three characters, an easier setting than full-code assignment, quantified in Table~\ref{tab:qa} and Appendix~\ref{app:qa}.

\emph{Rationale validity.} In a blinded clinician audit (Appendix~\ref{app:audit}), rationales from the verifier loop did not measurably improve on zero-shot rationales from the same generator; accepted and rejected rationales did not differ, and verifier score was uncorrelated with rated support. The measurable benefit of the gates is outcome alignment and ranking, not rationale quality.

\emph{Limitations of NormDot.} A faithful reading of an atypical presentation scores below a prototypical one, and it cannot distinguish a symptom the model missed from one the note never documented. Without a per-note symptom gold standard, extraction accuracy and the effect of a more precise symptom verifier remain unmeasured.

\emph{Reference dependence.} \wikidoc{} is a peer-edited wiki with incomplete coverage and editorial biases, and its fixed candidate lists cap recall at $0.886$ on our annotated frame (\S\ref{sec:annotation}). Whether the pipeline transfers to a formal ontology (SNOMED CT, UMLS) or a curated clinical reference (AMBOSS, UpToDate) remains untested. Requiring a \wikidoc{} match also selects the cohort: $60.7\%$ of candidate admissions are excluded because their discharge diagnosis has no usable \wikidoc{} row for the complaint (Appendix~\ref{app:mapping}). Excluded admissions lean toward symptom-coded, infectious and injury diagnoses and away from circulatory and respiratory ones, while demographics and length of stay barely differ. Our results therefore describe admissions whose diagnosis \wikidoc{} covers.

\emph{Scope of the comparisons.} The verifiers that accept training traces are the same metrics we report at test time, so a fine-tuned model is partly rewarded for conforming to the verifier rather than for diagnostic quality; ICD F1 against \mimic{} discharge codes is the one metric external to the \wikidoc{} pipeline, which is why we treat it as primary.  The encoder rows are retrained on our $10{,}400$-admission subset; full-data encoders score higher on macro-F1 (\S\ref{sec:results}), and the macro/micro trade-off means encoder rankings depend on the operating point.  Training and test also differ at the coding stage. During construction it conditions on the lab-reranked differential; at test time labs are not yet available, so it receives the diagnosis-stage output under a variant of the prompt that asks the model to reason about likely laboratory findings instead (Appendix~\ref{app:prompts}). An admission-only construction pipeline would remove this mismatch. We do not compare against outcome-only reasoning supervision such as DeepICD-R1.

\emph{Scope of the evidence.} Outside the 8B fine-tuned rows and the encoder row, the $\pm$ values in Table~\ref{tab:main} are evaluation repeats and understate true variance, holding the checkpoint fixed; only the 8B recipe and its label-only control were retrained across training seeds (\S\ref{sec:robustness}). Selecting checkpoints by ICD macro-F1 costs upstream output validity and top-1 ranking (\S\ref{sec:ranking}), and a different selection metric would move those columns. Reported clinician agreement ($\kappa=0.913$) covers the $236$ cases annotated by both of the first two reviewers; a third reviewer, who marked only the single most likely diagnosis, is excluded from it (Appendix~\ref{app:annotation}).  The admission-wise split permits patient overlap, which may affect individual predictions despite similar aggregate results on non-overlapping admissions (\S\ref{sec:error_shift}). The evaluation sampler applies no repetition penalty.

\emph{Task and label limitations.} Admission notes document findings but rarely justify them, and ICD codes are administrative labels shaped partly by billing practice rather than a clean record of the diagnostic process; both make the supervision signal noisier than purpose-built annotation would be.  In the error decomposition, the chronic-comorbidity class uses an Elixhauser-derived code set and the principal-diagnosis class assumes the first gold code is principal (Appendix~\ref{app:qa}); gold data also contains history and medication codes base models essentially never emit, so part of the base miss rate reflects label-space mismatch rather than clinical error. Finally, the dataset spans seven high-frequency chief complaints and should be extended to rarer ones and broader domains.

\section*{Ethical Considerations}
We use de-identified \mimic{} data under its data-use and privacy requirements. The data may reflect demographic skews and documentation practices specific to \mimic{}. \merlin{} is a research resource, not a deployed diagnostic tool, and must not inform clinical decisions without prospective validation and regulatory review.
\mimic{} is used under the PhysioNet Credentialed Health Data License 1.5.0;
\merlin{} is derived from it and will be released under the same terms,
requiring the same CITI training and credentialing.
The \wikidoc{} content used as an external reference is licensed under CC BY-SA 3.0 and is used and attributed under those terms.
\section*{Use of AI Assistants}
We used proprietary AI assistants to assist with revising prose, identifying inconsistencies, and debugging code. All suggestions and generated material were reviewed, verified, and edited by the authors, who take full responsibility for the paper's content, implementation, and results.

\section*{Author Contributions}
JF, DF, and CF contributed equally to this work through complementary scientific contributions; JF led the project. All three jointly conceived the project. \textbf{JF} led the design and implementation of the generator--verifier pipeline, the fine-tuning and evaluation framework, and the experimental program, conducted the computational experiments and analyses, produced the figures, and led the manuscript revisions. \textbf{DF} introduced the external clinical reference and built the \wikidoc{} reference corpus, designed the clinical data methodology and the clinician annotation protocol, co-developed the clinical data preprocessing with CF, contributed to early pipeline and fine-tuning development, led the clinical validation and the analysis of physician annotations, and coordinates the dataset release through PhysioNet. \textbf{CF} co-developed the clinical data preprocessing with DF, implemented the encoder baselines, coordinated the clinical collaboration, and prepared the initial manuscript draft. KS, RN, and JAC contributed clinical expertise and annotated the reference targets. JF, DF, and CF jointly discussed the methodology and results and contributed to the manuscript through collaborative writing and revision sessions. AL supervised the project with support from FG; AL and FG acquired the funding. Detailed CRediT roles are provided in Appendix~\ref{app:credit}.

\section*{Acknowledgments}
This work is funded by the Deutsche Forschungsgemeinschaft (DFG, German Research Foundation) Project-ID 528483508 -- FIP 12, as well as the European Union under the grant projects 101079894 (COMFORT -- Improving Urologic Cancer Care with Artificial Intelligence Solutions) and 101287679 (TWIN-X -- Multimodal Digital TWINs with Generative AI for eXplainable Precision Medicine). The views expressed are solely those of the authors and do not necessarily reflect those of the European Union or European Health and Digital Executive Agency (HaDEA); neither is responsible for them.

\bibliography{refs}

\appendix
\appendix

\section{Reference Targets and Clinician Annotation}
\label{app:grounding}

\subsection{Automatic Mapping Quality}
\label{app:mapping}
Each \mimic{} case is mapped to a \wikidoc{} disease row by the agreement of two
independent rankings of its discharge diagnosis: its position among ICD codes
matching a \wikidoc{} disease regex, and its position in an embedding-similarity
ranking over disease names (\texttt{neuml/pubmedbert-base-embeddings}).
Exact means both rank $0$, near means both within rank $1$, fuzzy means either at
rank $2$. Cases enter \merlin{} only if they match. We keep exact and near matches
for the two highest-volume complaints (abdominal pain, dyspnea) and additionally
admit fuzzy matches for the remaining five, where exact-plus-near leaves too few
cases. Unmatched and worse-than-fuzzy cases are dropped throughout.

\begin{figure}[htbp]
\centering
\includegraphics[width=0.99\columnwidth]{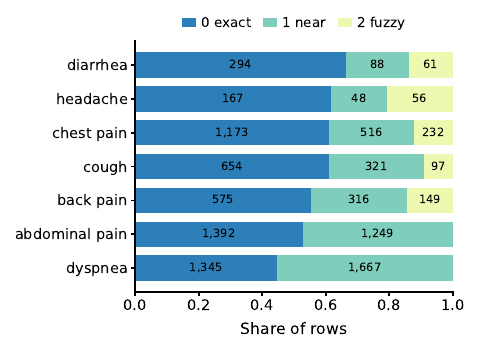}
\caption{\wikidoc{} mapping quality per chief complaint (share of rows: exact / near / fuzzy match to the disease row).}
\label{fig:match}
\end{figure}

Figure~\ref{fig:match} shows the composition of the released data per chief
complaint. It describes what \merlin{} contains rather than checking the mapper:
the $0\%$ fuzzy share for abdominal pain and dyspnea follows from the stricter rule
applied to them. Among the five complaints that admit fuzzy matches, the fuzzy share
stays under $15\%$ everywhere except headache ($20.7\%$). Headache is also the
smallest complaint (Table~\ref{tab:dataset}) and has the thinnest dual-annotation
coverage ($20$ of $50$ cases, Appendix~\ref{app:annotation}), so it is the least
independently checked.

\paragraph{Attrition from the mapping.}
The match requirement excludes $18{,}209$ of $30{,}011$ candidate admissions ($60.7\%$; Table~\ref{tab:attrition}). Across complaint--admission pairs, $12{,}331$ exclusions have a matching code ranked worse than fuzzy, $6{,}995$ have no discharge code in the complaint's \wikidoc{} list, and $1{,}139$ are fuzzy matches for the two complaints that admit only exact and near matches. Excluded and retained admissions are similar in sex ($56.2$ vs.\ $52.2\%$ female), median length of stay ($3.1$ vs.\ $3.3$ days), in-hospital death ($0.6$ vs.\ $0.3\%$) and median code count ($13$ vs.\ $12$). They differ in the primary diagnosis: excluded admissions more often carry a symptom code (ICD chapter XVIII, $12.1$ vs.\ $1.4\%$), an infectious ($8.9$ vs.\ $3.5\%$) or an injury diagnosis ($5.7$ vs.\ $2.3\%$), and less often a circulatory ($17.2$ vs.\ $31.5\%$) or respiratory one ($8.5$ vs.\ $19.8\%$). \merlin{} thus over-represents admissions with a specific diagnosis that \wikidoc{} lists for the complaint.

\begin{table}[htbp]
\centering
\small
\begin{tabular}{lrrr}
\toprule
\textbf{Complaint} & \textbf{Candidates} & \textbf{Retained} & \textbf{Excl.\ \%} \\
\midrule
Abdominal pain & $11{,}587$ & $3{,}316$ & $71.4$ \\
Back pain & $2{,}626$ & $1{,}063$ & $59.5$ \\
Chest pain & $5{,}095$ & $1{,}970$ & $61.3$ \\
Cough & $2{,}126$ & $1{,}110$ & $47.8$ \\
Diarrhea & $2{,}059$ & $448$ & $78.2$ \\
Dyspnea & $8{,}067$ & $4{,}239$ & $47.5$ \\
Headache & $1{,}322$ & $271$ & $79.5$ \\
\midrule
All pairs & $32{,}882$ & $12{,}417$ & $62.2$ \\
Unique admissions & $30{,}011$ & $11{,}802$ & $60.7$ \\
\bottomrule
\end{tabular}
\caption{\textbf{\wikidoc{} mapping attrition per chief complaint}, before the split and the one-complaint-per-subject assignment that yields the $10{,}400$ released admissions. An admission can be a candidate for several complaints, hence pairs vs.\ unique admissions.}
\label{tab:attrition}
\end{table}

\subsection{Clinician Annotation Protocol}
\label{app:annotation}
Three clinician co-authors independently review admission cases and confirm the
clinically correct diagnosis, given the case's chief complaint and the fixed
\wikidoc{} candidate list. The first completed the full stratified frame ($350$
cases, $50$ per chief complaint). The second annotated $236$ of them (abdominal,
back and chest pain, cough, diarrhea), on which candidate-level agreement is high
(Cohen's $\kappa=0.913$, exact-set agreement $89.0\%$, mean Jaccard $0.912$). The
third annotated $121$ cases (primarily dyspnea, diarrhea and headache), bringing annotation by at least two clinicians
to $316$ of $350$ cases across all seven complaints.

The third reviewer applied a narrower protocol, marking only the single most likely
diagnosis rather than every documented diagnosis that fits. Mean selections per case
are $1.11$, $1.06$ and $1.02$ for the three reviewers, and multi-label selection
rates are $9.7\%$, $5.5\%$ and $1.7\%$.  In $97$ of $121$ cases ($80.2\%$) the
third reviewer's selection is a subset of the first reviewer's, in $82$ cases
($67.8\%$) the two are identical, and only $2$ cases show strict reverse containment.
Typical divergences are cases such as a summary recording both ulcerative colitis
and CMV colitis, where the first reviewer marks both and the third marks the
principal one. Agreement is computed between the two reviewers who followed the same annotation protocol.

Where the ICD-derived and cosine-similarity references disagree on the confirmed
diagnosis (\S\ref{sec:annotation}), the clinician sided with the ICD-derived
reference roughly $2$:$1$ ($71$ vs.\ $39$ cases).

\section{Clinician Audit of Rationales}
\label{app:audit}
The audit asks whether the verifier loop produces better rationales, judged independently of the verifiers. Two clinician co-authors rated diagnosis-stage and coding-stage rationales from $42$ test admissions of the annotated frame (Appendix~\ref{app:annotation}) generated by Qwen3-32B, blinded to condition, verifier score and loop outcome, with each rationale shown next to its admission note. A rationale is the top-ranked target and its reason field from the generator's JSON answer, before reconciliation (Appendix~\ref{app:prompts}). Each admission contributes its in-loop rationale and a zero-shot rationale from the same generator, prompt and stage (the untuned test evaluation run of Table~\ref{tab:main}), so the primary contrast, loop against zero-shot, is within admission. Admissions were sampled by loop outcome (accepted or rejected: $31$ vs.\ $11$ at the diagnosis stage, $22$ vs.\ $20$ at the coding stage), so the accepted--rejected contrast is between admissions. For each rationale, raters answered whether its stated facts are supported by the patient information and whether they clinically support the stated target (Yes/No/Unclear; a rationale is valid only if both are Yes), and for each pair chose the clinically better rationale. The interface and questions were fixed before rating, and the analysis was implemented afterwards; the released protocol documents both and the deviations from an earlier pre-rating plan.

The loop did not improve validity: $-1.2$ points at both stages (95\% CIs $[-8.3, 7.1]$ and $[-16.7, 14.3]$). Accepted and rejected rationales differed by $-10.9$ to $+5.9$ points per rater and stage, with every CI spanning zero, and verifier score was uncorrelated with rated support (Spearman $-0.12$ to $+0.07$). Absolute levels differ by stage: diagnosis-stage rationales were judged valid in about $92\%$ and grounded in the note in at least $95\%$ of cases in both arms, while support for the specific ICD code at the coding stage was $67\%$ in both arms. Agreement on the coding-stage composite was $0.73$ ($\kappa=0.39$); diagnosis-stage $\kappa$ is deflated by the ceiling. The audit rates the diagnosis and coding stages only, not the symptom gate.

We release the annotation interface, protocol, rubric, analysis code, aggregate evaluation and agreement statistics, and synthetic demonstration cases at \url{https://github.com/DATEXIS/merlin-ddx-annotator}. Evaluated rationales and per-item aggregate ratings are distributed through PhysioNet under the \mimic{} data use agreement.

\section{Dataset Accounting}
\label{app:counts}
\begin{table}[t]
\centering
\small
\setlength{\tabcolsep}{5pt}
\begin{tabular}{lrrr}
\toprule
\textbf{Chief complaint} & \textbf{\#Cases} & \textbf{\#Sympt.} & \textbf{\#Diag.} \\
\midrule
Dyspnea      & 3{,}012 & 23 & 74 \\
Abdom.\ pain & 2{,}641 & 24 & 67 \\
Chest pain   & 1{,}921 & 27 & 48 \\
Cough        & 1{,}072 & 21 & 31 \\
Back pain    & 1{,}040 & 30 & 64 \\
Diarrhea     &   443 & 17 & 58 \\
Headache     &   271 & 13 & 19 \\
\midrule
\textbf{Overall} & \textbf{10{,}400} & \textbf{155} & \textbf{361} \\
\bottomrule
\end{tabular}
\caption{\merlin{} dataset composition. Each chief complaint maps to one \wikidoc{} table, from which \textbf{\#Sympt.} is the number of distinct symptom features that table tracks and \textbf{\#Diag.} the number of candidate diagnoses it lists; a single complaint therefore spans many of both, and the two counts are not comparable to \textbf{\#Cases}. The \textbf{Overall} row sums the per-table counts; tables share some features and diagnoses. The long-tail distribution of the discharge ICD labels is in Appendix~\ref{app:longtail} (Figure~\ref{fig:longtail_dist}, Table~\ref{tab:longtail}), alongside the head/body/tail breakdown it motivates.}
\label{tab:dataset}
\end{table}

Table~\ref{tab:counts} separates the levels at which \merlin{} can be counted,
since the headline figure ($124{,}510$) counts stage records and not accepted
traces. Each of the $10{,}400$ admissions is processed by all three generators,
giving $31{,}200$ four-stage chains and $124{,}800$ stage slots. Of these,
$124{,}510$ produced a stage record and $117{,}580$ carry a verifier score. Both
shortfalls sit almost entirely at the symptom stage, where output that does not
parse into the \wikidoc{} schema leaves no vector to score; the other stages lose
at most $657$ records each. Fine-tuning uses the train split only, so the headline
recipe trains on $82{,}789$ records: $55{,}974$ accepted traces and $26{,}815$
label-only fallbacks. The $4{,}799$ unscored symptom-stage train runs are dropped, because no symptom target can be rebuilt from the reference. Duplicating ICD-coding rows brings this to $104{,}686$
records, and the \texttt{drop} control removes the fallbacks instead, leaving
$55{,}974$.

\begin{table}[htbp]
\centering
\small
\setlength{\tabcolsep}{3.5pt}
\begin{tabular}{lrrrr}
\toprule
& \textbf{Sympt.} & \textbf{Diagn.} & \textbf{Lab} & \textbf{ICD} \\
\midrule
\multicolumn{5}{l}{\emph{All splits: $10{,}400$ adm., $31{,}200$ chains}}\\
Stage records & 30{,}910 & 31{,}200 & 31{,}200 & 31{,}200 \\
\quad scored & 24{,}782 & 31{,}198 & 30{,}543 & 31{,}057 \\
\midrule
\multicolumn{5}{l}{\emph{Train split: $7{,}299$ adm., $21{,}897$ chains}}\\
Scored records & 17{,}098 & 21{,}897 & 21{,}897 & 21{,}897 \\
\quad accepted & 12{,}918 & 18{,}813 & 15{,}566 & 8{,}677 \\
\quad fallback & 4{,}180 & 3{,}084 & 6{,}331 & 13{,}220 \\
\quad accept.\ rate & 75.6\% & 85.9\% & 71.1\% & 39.6\% \\
\bottomrule
\end{tabular}
\caption{\textbf{Dataset accounting by stage.} \emph{Stage records} are generated outputs; \emph{scored} are those a verifier could score; \emph{accepted} are those clearing the stage threshold, with the remainder replaced by the label-only \emph{fallback}. Row totals: $124{,}510$ stage records, $117{,}580$ scored, $82{,}789$ train-split records, $55{,}974$ accepted, $26{,}815$ fallbacks. The ICD stage accepts least often ($39.6\%$) and so does most of the filtering, consistent with its threshold sitting furthest above its stage average (\S\ref{sec:pipeline_stages}).}
\label{tab:counts}
\end{table}

\section{Pipeline Implementation Detail}
\label{app:pipeline_impl}
\begin{figure*}[t]
\centering
\includegraphics[width=0.94\textwidth]{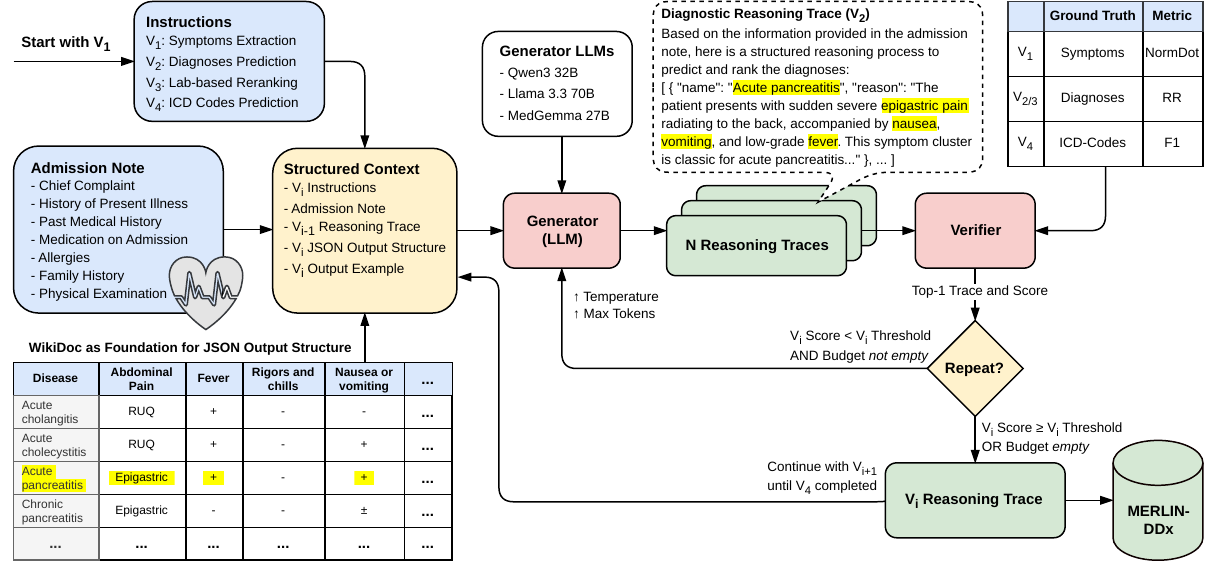}
\caption{\textbf{Implementation view of the \merlin{} pipeline}, expanding Figure~\ref{fig:generation}. Each stage assembles a structured context (the stage instruction, the admission note, the previous stage's accepted trace, and the stage's JSON output schema and example) and passes it to one of three generator LLMs, which samples $N$ candidate traces per round. The verifier scores each against that stage's reference target and keeps the top-1 (\wikidoc{} symptoms by NormDot for \Vone{}, \wikidoc{} differentials by RR for \Vtwo{}/\Vthree{}, \mimic{} discharge codes by F1 for \Vfour{}). Scores below threshold trigger regeneration at raised temperature and token budget; scores above threshold pass the trace on, while exhausting the budget replaces the trace with the label-only fallback. Lower left: the \wikidoc{} disease $\times$ manifestation table that defines the \Vone{} output schema and supplies the reference vectors (\texttt{+} present, \texttt{-} absent, \texttt{$\pm$} variable). Upper right: an excerpt of an accepted \Vtwo{} trace for an abdominal-pain case; highlighting marks manifestations shared between the trace's justification and the matched \wikidoc{} row. Both panels are illustrative rather than a verbatim record.}
\label{fig:pipeline_impl}
\end{figure*}

\begin{figure}[htbp]
\centering
\includegraphics[width=0.99\columnwidth]{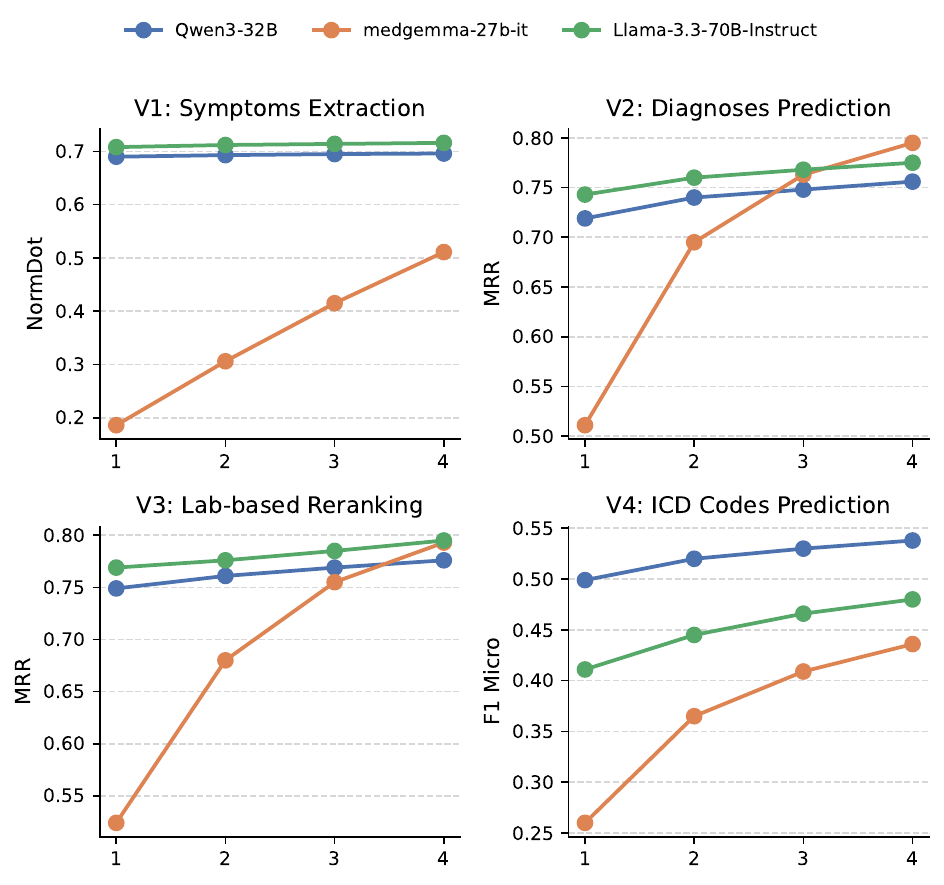}
\caption{Iterative improvement of verifier scores across generation-budget steps (abdominal-pain cases). As sub-threshold traces are regenerated, verifier scores rise consistently across pipeline stages.}
\label{fig:generation_plots}
\end{figure}

Figure~\ref{fig:pipeline_impl} expands Figure~\ref{fig:generation} into the form
actually run. It shows what the structured context handed to the generator
contains, what a \wikidoc{} reference row looks like as a table, and what an
accepted trace looks like as JSON. Figure~\ref{fig:generation_plots} shows verifier scores across generation rounds.

\section{Training, Checkpoint Selection and Robustness}
\label{app:training}
All SFT runs use AdamW, linear LR decay with $100$ warmup steps, weight decay
$0.01$, max sequence length $8{,}192$, bf16, gradient checkpointing, sequence
packing, $4$ epochs, and seed $42$. LoRA uses $\alpha=2r$ and dropout $0$. Full
fine-tuning of 14B/32B uses an FSDP path with sharded optimizer state, 8B uses
Unsloth DDP, and 0.6B uses Unsloth's offload path as a NaN workaround for tied
embeddings. Table~\ref{tab:training} lists per-run settings. Dataset construction samples $30$ candidates per round at temperature $0.4$ with $2{,}000$ output tokens ($2{,}200$ at the coding stage); each retry round raises the temperature by $0.2$ and the budget by $200$ tokens, for up to $4$ rounds. Downstream evaluation
serves each checkpoint with vLLM: every note runs once through symptom extraction
$\to$ diagnosis ranking $\to$ ICD coding, with lab reranking skipped, $4$
candidates per stage, temperature $0.4$, and the first valid JSON retained. The
evaluation repeats reported as $\pm$ in Table~\ref{tab:main} rerun this decoding
pipeline under seeds 43 and 44 against the same weights. Training used Unsloth
\texttt{2026.6.7} with \texttt{transformers 4.57.6}, \texttt{torch 2.10.0+cu128}
and CUDA 12.8; evaluation used vLLM~\citep{kwon_vllm_2023} \texttt{v0.17.1}. In total, the reported experiments used roughly $2{,}800$ GPU-hours on H200 and B200 GPUs: about $1{,}100$ for trace generation, $1{,}300$ for fine-tuning and $400$ for evaluation.

\paragraph{Checkpoints.} We refer to models by their Hugging Face identifier:
generators \texttt{Qwen/Qwen3-\{0.6B,8B,14B,32B\}},
\texttt{meta-llama/\allowbreak Llama-3.3-70B-Instruct} and
\texttt{google/medgemma-27b-it}; untuned baseline \texttt{baichuan-inc/\allowbreak Baichuan-M2-32B}; encoder baselines
\texttt{thomas-sounack/\allowbreak BioClinical-ModernBERT-base}
\citep{sounack2025bioclinicalmodernbertstateoftheartlongcontext}, \texttt{DATEXIS/sproto}
\citep{protopatient_van-aken-2022} and
\texttt{neuml/\allowbreak pubmedbert-\allowbreak base-embeddings}, a sentence-embedding model
derived from PubMedBERT \citep{gu2021pubmedbert} rather than that checkpoint
itself. The same embedding model supplies the similarity ranking in the \wikidoc{}
mapping (Appendix~\ref{app:mapping}).

\begin{table}[htbp]
\centering
\small
\setlength{\tabcolsep}{3.5pt}
\begin{tabular}{llccc}
\toprule
\textbf{Model} & \textbf{Mode} & \textbf{Hardware} & \textbf{Eff.\ bs} & \textbf{LR} \\
\midrule
0.6B & LoRA $r{=}128$ & 1$\times$H200 & 64  & $1\!\times\!10^{-4}$ \\
0.6B & Full          & 1$\times$B200 & 128 & $1\!\times\!10^{-4}$ \\
8B   & LoRA $r{=}128$ & 2$\times$B200 & 64  & $1\!\times\!10^{-4}$ \\
8B   & Full          & 2$\times$B200 & 256 & $1\!\times\!10^{-4}$ \\
14B  & LoRA $r{=}128$ & 2$\times$B200 & 64  & $5\!\times\!10^{-5}$ \\
14B  & Full (FSDP)    & 2$\times$B200 & 256 & $5\!\times\!10^{-5}$ \\
32B  & LoRA $r{=}256$ & 2$\times$B200 & 64  & $1\!\times\!10^{-4}$ \\
32B  & Full (FSDP)    & 4$\times$B200 & 256 & $1\!\times\!10^{-4}$ \\
\bottomrule
\end{tabular}
\caption{Per-run fine-tuning configuration.}
\label{tab:training}
\end{table}

\subsection{Checkpoint Selection Costs Upstream Quality}
\begin{figure*}[t]
\centering
\includegraphics[width=0.92\textwidth]{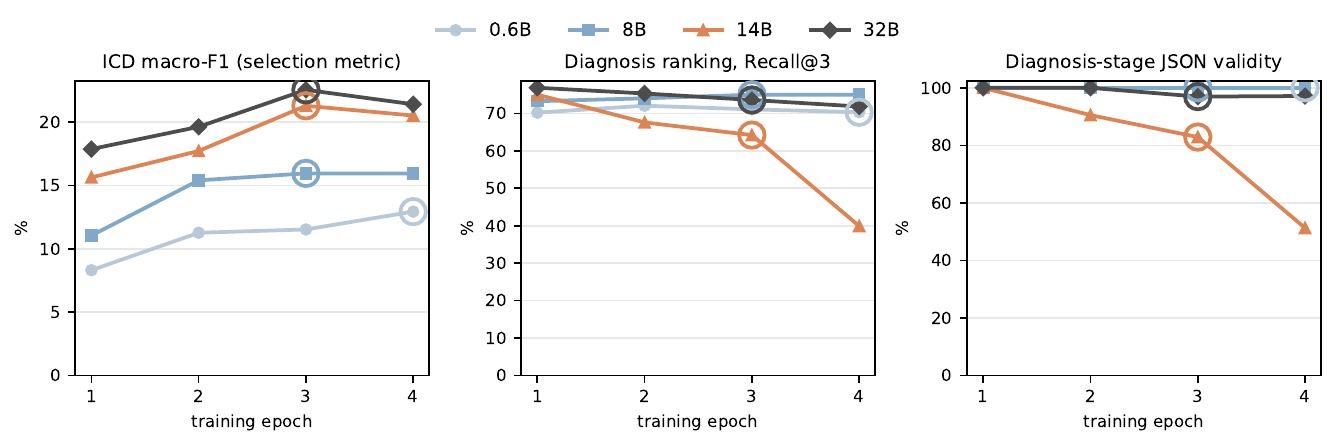}
\caption{\textbf{Selecting checkpoints by ICD macro-F1 costs upstream quality} (\S\ref{sec:ranking}). Dev-split epoch sweeps of the full fine-tuning runs, one line per model size; rings mark the epoch selected for Table~\ref{tab:main}. Left: the selection metric, which keeps rising into late epochs at every size. Middle and right: diagnosis Recall@3 and diagnosis-stage JSON validity, which do not. The trade is mild at 0.6B and 8B and severe at 14B, where the two right-hand panels track each other closely. A trace with unparseable diagnosis-stage output has no ranked list and scores zero on Recall@$k$.}
\label{fig:epoch_tradeoff}
\end{figure*}

The best epoch per model is selected on dev by ICD macro-F1.
Figure~\ref{fig:epoch_tradeoff} plots the full dev sweep behind
\S\ref{sec:ranking}. The 14B run is the extreme case: by epoch 4 it reaches
Recall@3 $39.9$, diagnosis validity $51.3\%$ and symptom validity $0.6\%$, well
past the epoch-3 checkpoint reported in the main text. Dividing Recall@3 by
diagnosis-stage validity gives $75.0$, $74.7$, $77.5$ and $77.9$ across epochs
1--4, so the ranking collapse tracks output format rather than diagnosis. At 8B the
sweep costs nothing measurable upstream (Recall@3 $73.3 \to 75.0$, validity
$100\%$), and at 32B Recall@3 falls only from $76.9$ to $73.6$.

\subsection{Training-Seed Robustness}
\begin{figure}[htbp]
\centering
\includegraphics[width=0.99\columnwidth]{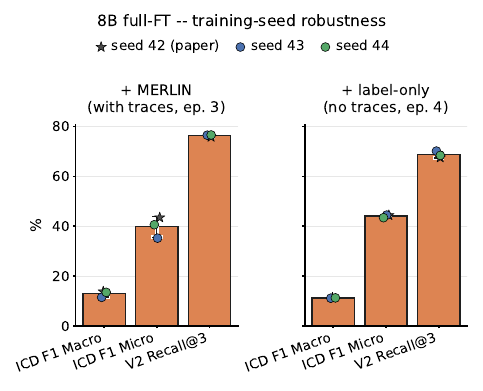}
\caption{Training-seed robustness for the 8B headline recipe (left, with reasoning traces) and its label-only control (right, no traces), each retrained end-to-end under seeds 42/43/44 and evaluated once on the test split, on a shared $y$-axis. Bars: mean $\pm$ std; markers: individual seeds (colored, star for seed 42); per-seed ICD macro-F1 for seeds 42/43/44 is $13.9$/$11.6$/$13.6$ with traces and $11.4$/$11.2$/$11.4$ label-only. The label-only control is far more stable on ICD F1 (macro CV ${\sim}1\%$ vs.\ ${\sim}10\%$ for the reasoning recipe) but less stable on diagnosis Recall@3 ($68.7{\pm}1.3$ vs.\ $76.3{\pm}0.4$); the ICD F1 micro gap between arms even points the opposite direction from macro on the seed-averaged mean, so the reasoning-supervision claim is scoped to macro-F1, not read as a uniform win.}
\label{fig:robustness}
\end{figure}

Figure~\ref{fig:robustness} retrains the 8B headline recipe and its label-only
control end-to-end under three seeds. 

\section{Additional Results}

\subsection{Per-Complaint Breakdown}
\label{app:cc}
\begin{table*}[t]
\centering
\small
\setlength{\tabcolsep}{4pt}
\begin{tabular}{lcccccc}
\toprule
\textbf{Chief complaint} & \textbf{n} & \textbf{NormDot} & \textbf{R@1} & \textbf{R@3} & $\mathbf{F1}_{\mathrm{mic}}$ & $\mathbf{F1}_{\mathrm{mac}}$ \\
\midrule
Dyspnea & 612 & $56.3$ & $31.5$ & $81.2$ & $53.6$ & $22.5$ \\
Abdominal Pain & 558 & $60.5$ & $40.9$ & $76.7$ & $49.0$ & $20.8$ \\
Chest Pain & 409 & $55.1$ & $33.5$ & $70.2$ & $52.2$ & $22.1$ \\
Back Pain & 244 & $78.5$ & $28.7$ & $63.5$ & $45.6$ & $21.7$ \\
Cough & 195 & $53.9$ & $36.4$ & $80.5$ & $48.4$ & $21.6$ \\
Diarrhea & 99 & $53.2$ & $38.4$ & $67.7$ & $43.9$ & $22.9$ \\
Headache & 67 & $53.2$ & $34.3$ & $73.1$ & $43.0$ & $20.6$ \\
\midrule
All & 2184 & $59.2$ & $34.8$ & $75.1$ & $50.5$ & $21.0$ \\
\bottomrule
\end{tabular}

\caption{\textbf{Chief-complaint breakdown} of the 32B\,+\,\merlinshort{} checkpoint (test split, $n{=}2{,}184$), same metrics as Table~\ref{tab:main}. Single evaluation seed ($42$), since the repeats in Table~\ref{tab:main} were not broken down by complaint, so the \textbf{All} row reads slightly differently from the three-repeat mean there. ICD macro-F1 is nearly flat across complaints ($20.6$--$22.9$) while the other columns vary far more (NormDot $53$--$79$, R@1 $29$--$41$, R@3 $64$--$81$), so long-tail coding difficulty does not track complaint-level symptom or diagnosis quality. Back pain's high NormDot ($78.5$) should be read with care: it has the longest symptom vector ($30$ dims against $13$--$27$ elsewhere), a scale effect not controlled for here, and it pairs with the lowest R@1 ($28.7$) and R@3 ($63.5$).}
\label{tab:cc-32b-full}
\end{table*}

Table~\ref{tab:cc-32b-full} breaks the 32B\,+\,\merlinshort{} row of
Table~\ref{tab:main} down by chief complaint. ICD macro-F1 is nearly flat across
complaints ($20.6$--$22.9$) while the upstream columns vary far more, so complaint
difficulty shows up before the ICD stage rather than in it.

\subsection{Symptom-Extraction Diagnostic}
\label{app:symptom}
NormDot values behind the symptom gate (seed-42 checkpoints, three evaluation repeats) are close to flat: $53.9 \to 59.8$ at 8B,
$59.1 \to 59.7$ at 32B, and a dip at 14B ($56.9 \to 52.3$). Because NormDot scores
agreement with a disease-level prototype rather than per-note extraction accuracy
(\S\ref{sec:pipeline_stages}), we do not read that flatness as evidence about
extraction quality, and it is not one of the evaluated tasks. Across training seeds
the gate is the noisiest quantity we measure ($57.9{\pm}4.6$ at 8B, against a CV
under $1\%$ for diagnosis Recall@3), and downstream ICD accuracy does not track it:
the seed with the highest NormDot ($61.8$) has the lowest ICD macro-F1 ($11.6$),
and the lowest-NormDot seed ($52.8$) is not the weakest on ICD ($13.6$). The gate
constrains which traces enter training without determining what the model does with
them.

\subsection{Encoder Baselines}
\label{app:encoders}
\begin{table}[t]
\centering
\small
\setlength{\tabcolsep}{4pt}
\begin{tabular}{lcc}
\toprule
\textbf{Encoder} & \textbf{F1$_{\mathrm{mic}}$} & \textbf{F1$_{\mathrm{mac}}$} \\
\midrule
\multicolumn{3}{l}{\emph{Matched data} ($10{,}400$ admissions, $3$ seeds)} \\
BioClin.\ ModernBERT & $26.4 \pm 0.8$          & $8.2 \pm 0.1$ \\
S-Proto              & $\mathbf{29.8 \pm 0.8}$ & $4.5 \pm 0.2$ \\
PubMedBERT           & $16.2 \pm 9.5$          & $4.3 \pm 1.5$ \\
\midrule
\multicolumn{3}{l}{\emph{Full \mimic{}} (\clinibench{})} \\
GatorTronS-345M      & $29.3$ & $\mathbf{25.5}$ \\
BioMedBERT-110M      & $24.5$ & $21.3$ \\
M2-80M               & $14.5$ & $10.8$ \\
\bottomrule
\end{tabular}
\caption{Encoder ICD-prediction baselines on the same \mimic{} subset (test split, $n{=}2{,}184$): mean $\pm$ std over $3$ training seeds, scored with the same metric as the ICD columns of Table~\ref{tab:main}. BioClinical ModernBERT has the best macro-F1 and is the row promoted into Table~\ref{tab:main}. S-Proto has the best micro-F1 ($29.8$) but a macro-F1 of only $4.5$: it wins on frequent codes and loses on the tail, the macro/micro trade discussed in \S\ref{sec:results}. All three trade precision for recall at the per-code level, over-predicting frequent codes rather than discriminating the tail; even the strongest reaches macro recall $13.8$ against macro precision $7.0$. PubMedBERT is the least seed-stable by a wide margin (micro-F1 std $9.5$ vs.\ $\le0.8$ for the other two). The \clinibench{} rows \citep{grundmann2025clinibench} are the full-\mimic{} ceiling, trained on $50$--$80\times$ more admissions under a top-20 ranked-list protocol rather than our threshold-based full-label prediction.}
\label{tab:encoders}
\end{table}

Table~\ref{tab:encoders} gives all three encoder baselines retrained on our
$10{,}400$-admission subset, plus the \clinibench{} full-\mimic{} rows scored on our
test admissions. BioClinical ModernBERT has the best macro-F1 and is the row
promoted into Table~\ref{tab:main}. Its $8.2$ macro-F1 beats untuned MedGemma-27B
($7.6$) and the untuned Qwen3-8B base ($5.0$), so a small supervised classifier is
stronger than an untuned generative model on long-tail ICD.
\merlinshort{} exceeds the encoder at every size, and the label-only controls (Table~\ref{tab:main}) isolate how much of that comes from the traces. One of PubMedBERT's three seeds (recall $96.6\%$,
precision $2.7\%$) behaves like an all-positive predictor; the mean includes all three seeds.

\subsection{Head, Body and Tail Detail}
\label{app:longtail}
\begin{table}[t]
\centering
\small
\setlength{\tabcolsep}{6pt}
\begin{tabular}{lccc}
\toprule
\textbf{Size} & \textbf{Head} & \textbf{Body} & \textbf{Tail} \\
\emph{\#labels} & \emph{153} & \emph{349} & \emph{328} \\
\midrule
0.6B base & $0.3$ & $0.1$ & $0.0$ \\
0.6B $+$\,\merlinshort{} & $31.6$ & $9.4$ & $0.6$ \\
\addlinespace[1pt]
8B base & $12.4$ & $5.5$ & $0.9$ \\
8B $+$\,\merlinshort{} & $32.4$ & $15.3$ & $3.6$ \\
\addlinespace[1pt]
14B base & $24.3$ & $12.0$ & $3.6$ \\
14B $+$\,\merlinshort{} & \underline{$37.5$} & \underline{$21.4$} & \underline{$7.5$} \\
\addlinespace[1pt]
32B base & $26.3$ & $13.2$ & $5.7$ \\
32B $+$\,\merlinshort{} & $\mathbf{40.1}$ & $\mathbf{23.3}$ & $\mathbf{9.6}$ \\
\bottomrule
\end{tabular}

\caption{\textbf{Head, body and tail, full detail.} Companion to the ``F1 by train.\ frequency'' block of Table~\ref{tab:qa} (\S\ref{sec:longtail}), with the same test-split F1 numbers at every model size plus \emph{\#labels}, the count of each stratum's codes present in the test gold. Strata are three-character codes grouped by \merlin{} training frequency: \emph{head} $\ge100$ admissions ($153$ codes, $82.8\%$ of all training code occurrences), \emph{body} $10$--$99$ ($15.0\%$), \emph{tail} $<10$ including codes never seen ($2.1\%$). These $830$ gold-present codes are the macro-F1 universe of Table~\ref{tab:main}: predicted codes outside it are discarded before scoring, so the strata average back, weighted by \#labels, to the macro-F1 of the same seed-42 evaluation runs (Table~\ref{tab:lora}). Counting discarded codes as false positives over the union of gold and predicted codes instead lowers macro-F1 by up to $3.1$ points, most for the larger untuned models, and narrows the \merlinshort{}--label-only gap of the seed-42 runs from $2.5$--$6.5$ to $1.5$--$6.0$ points without reversing it at any scale.}
\label{tab:longtail}
\end{table}

\begin{figure}[t]
\centering
\includegraphics[width=0.98\linewidth]{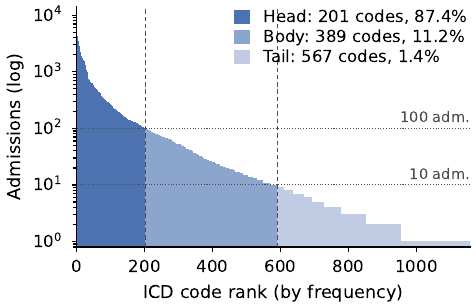}
\caption{\textbf{Label-frequency distribution.} Three-character ICD discharge codes across all $10{,}400$ admissions, rank-ordered, log-scaled. Shading applies the $\ge100$ / $10$--$99$ / $<10$ admission thresholds: $201$ head codes carry $87.4\%$ of the $123{,}133$ admission--code pairs, while $567$ tail codes share $1.4\%$. Table~\ref{tab:longtail} applies the same thresholds to \emph{training} frequency alone, giving $153$/$359$/$583$ codes over the full label space, of which $153$/$349$/$328$ occur in the test gold labels that Tables~\ref{tab:qa} and~\ref{tab:longtail} score.}
\label{fig:longtail_dist}
\end{figure}

Table~\ref{tab:longtail} gives the per-size values behind the frequency block of
Table~\ref{tab:qa}, plus the number of gold codes in each stratum.
Figure~\ref{fig:longtail_dist} shows the label-frequency distribution over the
full dataset; the strata above cut the same thresholds on training frequency
alone. The improvement over the untuned base remains under full-code evaluation (Table~\ref{tab:qa}, last column), but from 8B the label-only control is ahead at this granularity, and the absolute rate stays under $6$ macro-F1 everywhere: models reach the right category far more often than the exact code a coder assigns.

\subsection{Differential Length}
Predicted differentials converge to roughly the same length after fine-tuning
($7.4\to9.9$ at 32B, $3.5\to10.0$ at 0.6B, $3.8\to9.8$ at 8B). The 32B base starts
nearly twice as long as the smaller bases, leaving less headroom before the list
saturates. This is the mechanism we appeal to in \S\ref{sec:ranking} for the
Recall@1/Recall@3 divergence at that scale.

\subsection{Reasoning-Free Rates}
\label{app:shortcut}
The fine-tuned models do not always write out their reasoning. We call a valid generation \emph{reasoning-free} when fewer than ten words precede the templated summary that closes each training trace's reasoning block or, without a summary, the JSON answer (Appendix~\ref{app:prompts}). At the coding stage, the \merlinshort{} checkpoints of Table~\ref{tab:main} are reasoning-free on $72$--$100\%$ of valid test outputs, whereas the untuned bases never are and the diagnosis stage mostly keeps its reasoning (Table~\ref{tab:shortcut}). We hypothesize that the coding-stage targets do not tie the reasoning to the answer: fallback rows carry empty reasoning blocks, which suppress Qwen3's reasoning \citep{twist_collapse_2026}, and accepted traces end in a summary that differs from the reconciled JSON (\S\ref{sec:dataset}). The 8B epoch-1 checkpoint still reasons but reaches only $11.1$ dev macro-F1, against $12.5$ for the best label-only checkpoint.
\begin{table}[t]
\centering
\small
\setlength{\tabcolsep}{4pt}
\begin{tabular}{@{}lrr@{}}
\toprule
\textbf{Checkpoint} & \textbf{Diag.\ ranking} & \textbf{ICD coding} \\
\midrule
\multicolumn{3}{@{}l}{\emph{Dev split, 8B data-construction recipes}} \\
No thresholds $+$ ICD$\times$2 & $0$ & $2$ \\
Full thresholds & $3$ & $14$ \\
Half thresholds & $100$ & $100$ \\
Full thresholds $+$ ICD$\times$2 & $1$ & $100$ \\
\midrule
\multicolumn{3}{@{}l}{\emph{Test split, checkpoints of Table~\ref{tab:main}}} \\
Untuned bases (all sizes) & $0$ & $0$ \\
0.6B $+$\,\merlinshort{} & $5$ & $72$ \\
8B $+$\,\merlinshort{} & $1$ & $100$ \\
14B $+$\,\merlinshort{} & $0$ & $93$ \\
32B $+$\,\merlinshort{} & $32$ & $100$ \\
\bottomrule
\end{tabular}
\caption{\textbf{Reasoning-free rate (\%) at the diagnosis-ranking and ICD-coding stages.} A valid generation is reasoning-free when fewer than ten words of reasoning precede the templated summary that closes every training trace's reasoning block or, if no summary is written, the JSON answer; outputs with invalid JSON are excluded from the denominator. Two routes reach it: writing only the summary (nearly all cases at 8B--32B and in the dev recipes) or answering straight into the JSON (all cases at 0.6B). Top: 8B data-construction recipes on the dev split, at their dev-selected epochs; full thresholds $+$ ICD$\times$2 is the \merlinshort{} recipe. Bottom: the checkpoints of Table~\ref{tab:main} on the test split (seed-42 evaluation run). Label-only controls are not shown: their text before the JSON is mostly echoed prompt sections, which word counts cannot tell from reasoning.}
\label{tab:shortcut}
\end{table}

\section{Data Construction and Adaptation Method}

\subsection{Data-Construction Ablation: Exact Values}
\label{app:ablation}
\begin{table*}[t]
\centering
\small
\setlength{\tabcolsep}{4.5pt}
\begin{tabular}{lccccccc}
\toprule
 & \textbf{Symptoms} & \multicolumn{2}{c}{\textbf{Diagnosis ranking}} & \multicolumn{4}{c}{\textbf{ICD codes}} \\
\cmidrule(lr){2-2} \cmidrule(lr){3-4} \cmidrule(lr){5-8}
\textbf{Training data} & \textbf{NormDot} & \textbf{R@1} & \textbf{R@3} & $\mathbf{F1}_{\mathrm{mic}}$ & $\mathbf{F1}_{\mathrm{mac}}$ & $\mathbf{R}_{\mathrm{mac}}$ & $\mathbf{P}_{\mathrm{mac}}$ \\
\midrule
8B base (untuned) & $53.6$ & $45.5$ & $64.7$ & $19.2$ & $5.5$ & $5.3$ & $9.8$ \\
\midrule
\multicolumn{8}{l}{\emph{No reasoning traces (label-only)}}\\
\quad label-only & $56.5$ & $\mathbf{53.3}$ & $72.3$ & $38.1$ & $9.5$ & $9.8$ & $11.8$ \\
\quad label-only $+$ ICD$\times$2 & $58.9$ & $49.8$ & $68.3$ & $\mathbf{43.8}$ & $12.5$ & $12.1$ & $15.4$ \\
\multicolumn{8}{l}{\emph{Reasoning traces, unfiltered}}\\
\quad all traces$ $ & $60.0$ & $45.7$ & $75.1$ & $32.7$ & $10.3$ & $11.0$ & $14.4$ \\
\quad all traces $+$ ICD$\times$2 & \underline{$61.2$} & $51.5$ & $74.8$ & $33.2$ & \underline{$13.1$} & \underline{$14.5$} & $17.5$ \\
\multicolumn{8}{l}{\emph{Reasoning traces, verifier-filtered}}\\
\quad $\geq\tfrac{1}{2}$ thresholds & $54.0$ & \underline{$52.9$} & $\mathbf{76.8}$ & $34.4$ & $11.2$ & $11.5$ & $16.1$ \\
\quad $\geq$\,thresholds & $\mathbf{61.4}$ & $52.0$ & \underline{$75.7$} & $34.8$ & $12.9$ & $13.9$ & \underline{$18.1$} \\
\quad $\geq$\,thresholds, drop not fallback & $\mathbf{61.4}$ & $50.0$ & $75.6$ & $33.0$ & $10.4$ & $11.2$ & $14.4$ \\
\quad $\geq$\,thresholds $+$ ICD$\times$2 (\merlinshort{}) & $58.5$ & $52.1$ & $75.0$ & \underline{$43.2$} & $\mathbf{16.0}$ & $\mathbf{15.8}$ & $\mathbf{20.1}$ \\
\bottomrule
\end{tabular}
\caption{\textbf{Data-construction ablation, exact values} (Qwen3-8B, full fine-tuning, dev split, dev-selected best epoch), plotted in Figure~\ref{fig:ablation}. Eight variants span two construction factors, how traces are filtered (row groups) and whether ICD-coding rows are duplicated ($+$\,ICD$\times$2), plus one drop-vs-fallback control. ICD columns are micro- and macro-F1 over three-character codes, then macro-averaged recall and precision. }
\label{tab:ablation}
\end{table*}

Table~\ref{tab:ablation} gives the values behind Figure~\ref{fig:ablation}, plus
the two controls the markers do not separate visually. Diagnosis ranking is included
to show that the construction choice that most improves ICD macro-F1 does not come
at its expense: Recall@3 stays within two points of the best variant across every
trace-based recipe. The \textbf{NormDot} column is the symptom gate
(\S\ref{app:symptom}), shown only to confirm the variants do not differ through it.

 Because the grid is not fully crossed, the
monotone filter-strictness trend in \S\ref{sec:ablations} is read within each oversampling setting only.

\subsection{Parameter-Efficient Adaptation (LoRA)}
\label{app:lora}
\begin{table*}[t]
\centering
\small
\setlength{\tabcolsep}{3pt}
\begin{tabular}{lccccccc}
\toprule
 & \textbf{Sympt.} & \multicolumn{2}{c}{\textbf{Diagn.}} & \multicolumn{4}{c}{\textbf{ICD codes}} \\
\cmidrule(lr){2-2} \cmidrule(lr){3-4} \cmidrule(lr){5-8}
\textbf{Model} & \textbf{NormDot} & \textbf{R@1} & \textbf{R@3} & $\mathbf{F1}_{\mathrm{mic}}$ & $\mathbf{F1}_{\mathrm{mac}}$ & $\mathbf{R}_{\mathrm{mac}}$ & $\mathbf{P}_{\mathrm{mac}}$ \\
\midrule
Qwen3-0.6B & $51.3$ & $34.4$ & $53.2$ & $1.2$ & $0.1$ & $0.1$ & $0.4$ \\
\quad$+$\,\merlinshort{} (full) & $\mathbf{61.6}$ & $51.1$ & $73.8$ & $43.7$ & $10.0$ & $9.9$ & $11.8$ \\
\quad$+$\,\merlinshort{} (LoRA $r{=}128$) & \underline{$61.4$} & \underline{$52.5$} & $74.2$ & $43.2$ & $9.8$ & $9.9$ & $11.5$ \\
Qwen3-8B & $53.9$ & $45.6$ & $67.1$ & $19.5$ & $4.9$ & $4.7$ & $8.9$ \\
\quad$+$\,\merlinshort{} (full) & $59.1$ & $\mathbf{53.4}$ & \underline{$75.8$} & $43.5$ & $13.9$ & $13.5$ & $17.6$ \\
\quad$+$\,\merlinshort{} (LoRA $r{=}128$)$^{\ast}$ & $53.1$ & $46.2$ & $63.7$ & $17.9$ & $4.9$ & $3.9$ & $12.2$ \\
Qwen3-14B & $56.9$ & $48.8$ & $68.5$ & $30.4$ & $10.9$ & $13.7$ & $14.4$ \\
\quad$+$\,\merlinshort{} (full) & $50.4$ & $45.4$ & $64.9$ & $49.0$ & $18.9$ & $18.6$ & \underline{$22.3$} \\
\quad$+$\,\merlinshort{} (LoRA $r{=}128$) & $61.0$ & $45.1$ & $75.7$ & $47.5$ & $17.0$ & $17.7$ & $19.2$ \\
Qwen3-32B & $59.0$ & $50.3$ & $70.6$ & $33.5$ & $12.7$ & $15.4$ & $15.7$ \\
\quad$+$\,\merlinshort{} (full) & $59.2$ & $34.8$ & $75.1$ & $\mathbf{50.5}$ & $\mathbf{21.0}$ & \underline{$21.1$} & $\mathbf{23.6}$ \\
\quad$+$\,\merlinshort{} (LoRA $r{=}256$) & \underline{$61.4$} & $37.7$ & $\mathbf{78.1}$ & \underline{$49.6$} & \underline{$20.6$} & $\mathbf{21.5}$ & $21.5$ \\
\bottomrule
\end{tabular}

\caption{LoRA vs.\ full fine-tuning and the untuned base, held-out test split, columns as in Table~\ref{tab:main}. Single evaluation run per model, so no $\pm$ is given. $^{\ast}$The 8B $r{=}128$ adapter stays at base-level ICD macro-F1 with $70\%$ coding-stage JSON validity; $r{=}256$ was lower at every epoch.}
\label{tab:lora}
\end{table*}

Table~\ref{tab:lora} reports LoRA as an adaptation-method ablation. At 0.6B, 14B and 32B, LoRA stays within $1$--$3$ points of full
fine-tuning on ICD F1 and matches or exceeds it on diagnosis Recall@3. The
deterministic error decomposition (\S\ref{sec:error_shift}) confirms it reproduces
the full fine-tuning error profile and not just the aggregate score: at 32B,
macro-F1 $21.0$ vs.\ $20.6$, Primary $67.5$ vs.\ $68.9$, Hist.\ F1 $53.1$ vs.\ $50.7$.

At 8B the $r{=}128$ adapter reaches a test ICD macro-F1 of $4.9$, the untuned base's level, with JSON validity of $70\%$ at the coding stage and $33\%$ at symptom extraction; $r{=}256$ is lower on both. 

\section{Error Decomposition Detail}
\label{app:qa}

\paragraph{Class definitions.}
All classes in Table~\ref{tab:qa} are set operations over gold and predicted
three-character ICD categories on the test split ($n{=}2{,}184$). A predicted code
\emph{covers} a gold code when they share a three-character category. \textbf{Primary} is the share of cases whose predictions cover the first gold code, which we take as the principal diagnosis (see caveats). \textbf{Hist., Med.\ and Chron.\ F1} restrict gold and predicted codes to one class, pool true positives, false positives (predicted class codes not in the gold set) and false negatives over all test cases, and report the resulting micro-F1. The classes are Z80--Z99 for history, Z79 for medication, and an Elixhauser-derived chronic set: CHF/IHD (I50, I11,
I13, I09, I42, I43, I25), arrhythmia (I44--I49), valvular (I05--I08, I34--I39),
pulmonary circulation (I26--I28), peripheral vascular (I70, I71, I73), hypertension
(I10, I12, I15), chronic neurological (G20, G35, G40, G41, G81--G83), chronic
pulmonary (J40--J47, J60--J67), diabetes (E10--E14), hypothyroidism (E00--E03, E89),
renal (N18, N19, N25), chronic liver (K70--K74, K76, B18, I85), peptic ulcer
(K25--K28), HIV (B20--B24), malignancy (C00--C96), rheumatologic (M05, M06, M08,
M30--M36, L94), coagulopathy (D65--D69), obesity (E66), chronic anemia (D50--D53,
D63, D64), substance use (F10--F19), and chronic psychiatric (F20--F34). Acute
categories are deliberately excluded.

\paragraph{Full code.} The rightmost column of Table~\ref{tab:qa} repeats the
base-vs-\merlinshort{} comparison at exact ICD granularity (\texttt{K5660}, not the
three-character \texttt{K56} used elsewhere). Predicted codes are read from the raw
ICD-stage output, normalized (case, punctuation, comma- and slash-separated cells
split) and validated against a single-code ICD-10-CM pattern; malformed tokens are
dropped. We report macro- rather than micro-F1 because full code spans the entire
label space, where the micro/macro gap is largest. Full-code macro-F1 rises with
fine-tuning at every scale ($0.0\to1.5$ at 0.6B, $1.4\to3.5$ at 32B) but stays under
$4$ for the base and \merlinshort{} rows and reaches $5.4$ for the 32B label-only control, against $20.9$ for the same 32B checkpoint at category level
(Table~\ref{tab:main}) and $26.6$ for full-code micro-F1. Part of the gap to the label-only control is built into the training targets: reconciliation matches codes at the category level and keeps the generator's own full code (Appendix~\ref{app:prompts}), so \merlinshort{} is supervised on generator-chosen subcodes where the label-only control sees the discharge codes. The full-code comparison is therefore tilted towards the label-only control.

\paragraph{Macro-F1 without history, medication and chronic codes (\S\ref{sec:error_shift}).} History, medication and
chronic-comorbidity codes are arguably easier to recover than an acute diagnosis,
since they are often stated verbatim rather than inferred. We exclude these three
families from both gold and predicted code sets, recompute macro-F1, and rerun the
case-level paired bootstrap ($2{,}000$ resamples, seed $42$) over the same $2{,}184$
test cases. $53\%$ of gold code occurrences remain. The gain survives: base $3.5 \to$
\merlinshort{} $10.3$ ($\Delta{=}{+}6.8$, 95\% CI $[{+}5.8, {+}7.1]$), against
$\Delta{=}{+}8.9$ on all codes.

\paragraph{Memorization within the overlapping admissions.} Restricted to the $390$
test admissions whose patient also appears in train (\S\ref{sec:error_shift}), codes
that also occur in that patient's own training admission are recovered far more
often than codes gold in the test admission but never seen for that patient: $54.1$
vs.\ $30.7$ recall for personal history and $63.6$ vs.\ $50.0$ for long-term
medication, alongside the chronic-comorbidity pair reported in the main text.

\paragraph{Caveats.}
The principal-diagnosis proxy assumes gold ordering preserves \texttt{seq\_num},
which we did not verify against the discharge-note diagnosis section, so that column
is indicative while the history and medication columns are exact. The chronic set
above is one defensible mapping among several. 

\section{Pipeline Prompts}
\label{app:prompts}
Each pipeline stage is driven by a single generation prompt, checked by its
verifier against WIKIDOC-derived targets (\Vone{}--\Vthree{}) or the recorded
MIMIC-IV discharge codes (\Vfour{}); \Vthree{} reranks \Vtwo{}'s output once
laboratory results become available and shares \Vtwo{}'s diagnosis-ranking target.
We reproduce one representative prompt per verifier below. \Vone{} is the generic
extraction template, shared by six of the seven chief complaints; abdominal pain
uses a bespoke variant that inlines its manifestation definitions. Curly-brace
fields (e.g.\ \texttt{\{admission\_note\}}) are per-case slots. The full prompt set
ships in the repository linked in \S\ref{sec:intro}: the abdominal-pain \Vone{}
variant, the label-only baseline prompts of \S\ref{sec:ablations}, and the
admission-time evaluation prompts that replace \Vone{}--\Vfour{} at test time. The
reproductions below silently correct two spelling errors; the released templates
are byte-identical to what generated the dataset.
\paragraph{From completion to training target.} Each prompt asks for a plain-text candidate list and
rerank before the final JSON object. The extractor returns the \emph{last}
schema-valid object in the completion, and the verifier scores that object; the JSON-validity rates in \S\ref{sec:dynamics} are measured on that parse. Everything before the closing think tag or the first code fence becomes the reasoning block of the training target.

Two further steps shape every accepted trace that scores below $1$. First, its answer is reconciled with the reference. At the coding stage, predicted codes whose three-character category is absent from the discharge codes are removed and discharge codes from missing categories are added with a placeholder reason (\emph{``Based on the admission note code X is suspected.''}); retained codes keep the generator's full code, which can differ from the discharge code below the category level (gold \texttt{E11.9}, retained \texttt{E11.65}); at the diagnosis and reranking stages, the confirmed diagnosis is moved to rank~1, or inserted there with a placeholder reason. Of accepted coding-stage traces, $99.2$--$99.8\%$ are reconciled per generator, $34$--$40\%$ of their reason fields are placeholders, and $91$--$92\%$ contain at least one. Second, a templated summary closes the reasoning block. It lists the model's own, unreconciled answer and hands over to the reconciled JSON (coding stage: \emph{``I think the patient has the following ICD codes: 1:~\dots{} After a final consideration, I think the patient has the following ICD codes:''}; the diagnosis stages use the analogous wording). A reasoning block can therefore end in a list that differs from the answer that follows it. Fallback rows, whose traces never cleared the threshold, keep the reconciled model answer, or the reference answer if no valid JSON exists, under an empty reasoning block; symptom-stage runs without valid JSON are dropped (Appendix~\ref{app:counts}).

\paragraph{\Vone{}: symptom extraction.}\leavevmode\par\nobreak
\begin{lstlisting}
Your task is to extract and classify all relevant symptoms and clinical manifestations
from a patient's admission note, then summarize them in a single valid JSON object.

Respond clearly and concisely. Avoid speculation, filler, or disclaimers.

### Required Output:

1. A brief reasoning process (internal steps to identify manifestations).
2. A list of present manifestations and a list of absent manifestations.
3. A single JSON object containing all manifestations with reasoning.
4. Stop after printing the JSON object - no extra commentary.

### Reasoning Workflow:

1. Task Familiarization: View Annotation Rules.
2. Review Note: Read the admission note and identify all manifestations mentioned.
3. Create Lists: Make two lists - one for present manifestations, one for absent manifestations.
4. Check Completeness: Ensure all manifestations are correctly classified. Revise if needed.
5. Generate JSON: Output one valid JSON object that follows the specified JSON structure / Output Format Example.

### Annotation Rules:

* Values:
  - 1: Present - clearly documented at admission.
  - -1: Absent - explicitly denied, ruled out, or resolved prior to admission.
  - 0: Unmentioned - no reliable information in the note.

* Temporal Scope:
  - Annotate only symptoms present at admission.
  - Assign 0 if timing is unclear.
  - Exclude resolved or prior symptoms.

* Objective vs. Subjective Evidence:
  - Record objective signs only if measured/observed.
  - Record subjective symptoms if clearly reported by the patient.

### Admission Note
{admission_note}

### JSON structure
Return your output as one valid JSON object.
Each key is the manifestation name and each value is an object with:
* "value": one of 1, -1, or 0
* "reasoning": 1 to 3 sentences only of how the evidence supports the assigned value; required only if value is 1 or -1, otherwise "None"

### Output Format Example
{symptom_format_example}
\end{lstlisting}

\paragraph{\Vtwo{}: diagnosis ranking.}\leavevmode\par\nobreak
\begin{lstlisting}
Your task is to predict diagnoses and print them once in valid JSON.
Respond clearly and concisely. Avoid speculation, filler, or disclaimers.

### Required Output:

1. A brief reasoning process (internal steps to identify diagnoses).
2. A list of potential diagnoses.
3. A single JSON object containing all diagnoses with reasoning.
4. Stop after printing the JSON object - no extra commentary.

### Structured Reasoning Process

1. Analyze Patient Data: Review the Admission Note and Clinical Manifestations.
2. Analyze Potential Diagnoses: Review Potential Diagnoses and think about their likelihood.
3. Generate Predictions:
  - Select and rank the 10 most likely diagnoses.
  - Print them as a simple list, NOT in JSON Format.
4. Refine & Rerank:
  - How are the Manifestations related to your diagnoses?
  - Is every given name one of the potential diagnoses?
  - Rerank diagnoses accordingly. NOT in JSON Format.
5. Output JSON Once: Provide your final ranked diagnoses and reasoning in JSON.
  - Use the format of the Output Format Example.
  - Include only a short reasoning of 1 - 3 sentences for each diagnosis.
  - The JSON is the only output. Stop after printing it.

### Admission Note
{admission_note}

### Clinical Manifestations
{clinical_manifestations}

### Potential Diagnoses
{potential_diagnoses}

### Output Format Example
{diagnose_format_example}
\end{lstlisting}

\paragraph{\Vthree{}: lab reranking (generation only).} \Vthree{} is the \Vtwo{} template with three changes, so we describe it rather than reproduce it. Its preamble states that the ranked list was produced without laboratory results and that results at admission time are now available; a \texttt{\{laboratory\_results\}} block is inserted before \texttt{\{potential\_diagnoses\}}; and steps 2--3 of the reasoning process ask how the laboratory results affect each diagnosis's likelihood and request a \emph{reranked} rather than a fresh list. The output schema and format example are unchanged. The verbatim template ships in the repository. At evaluation the \Vfour{} prompt instead asks the model to reason about \emph{likely} laboratory findings it does not have.

\paragraph{\Vfour{}: ICD prediction.}\leavevmode\par\nobreak
\begin{lstlisting}
Your task is to predict and rank ICD-10 discharge codes based on admission information.
Respond clearly and concisely. Avoid speculation, filler, or disclaimers.

### Required Output:

1. A brief reasoning process (internal steps to identify diagnoses).
2. A list of potential ICD-Codes.
3. A single JSON object containing all ICD-Codes with reasoning.
4. Stop after printing the JSON object - no extra commentary.

### Structured Reasoning Process
1. Analyze Patient Data:
  - Admission Note
  - Laboratory Results at Admission Time
  - Clinical Manifestations
  - Potential Diagnoses
2. Predict many ICD-10 codes:
  - Predict up to 30 icd-codes based on 1. and put them in order of likelihood.
  - Think about all ICD-10 codes the patient could be assigned when leaving the hospital.
  - Go broad and prefer to predict more we can remove unlikely codes later.
  - Print the ICD-10 as a simple list, not in JSON format.
3. Rethink your predicted ICD-10 codes:
  - How are the Manifestations related to your diagnoses?
  - Which ones do not fit, which did you miss?
  - Is every ICD-10 code an official ICD-10 code?
  - Is the amount of ICD-10 codes between 3 and 30?
4. Print your final answer in json format.
  - Rerank the order if needed.
  - Use the format of the Output Format Example.
  - Give a reason for each code, why it is likely to be related to the patient.
  - The json can have 3 - 30 ICD-10 codes.
  - Stop after you printed the json, DO NOT print any other text or JSON.

### Admission Note
{admission_note}

### Laboratory Results at Admission Time
{laboratory_results}

### Clinical Manifestations
{clinical_manifestations}

### Potential Diagnoses
{potential_diagnoses}

### Output Format Example
{icd_format_example}
\end{lstlisting}

\section{CRediT Author Contributions}
\label{app:credit}
Roles follow the CRediT taxonomy (L = lead, E = equal, S = supporting).
\begin{table}[h]
\centering
\small
\begin{tabular}{lccc}
\toprule
Role & JF & DF & CF \\
\midrule
Conceptualization & E & E & E \\
Methodology & E & E & S \\
Software & L & S & S \\
Data curation & S & E & E \\
Investigation & L & S & S \\
Formal analysis & L & S & -- \\
Validation & E & E & S \\
Visualization & L & -- & -- \\
Resources & -- & L & -- \\
Project administration & L & S & S \\
Writing -- original draft & E & S & E \\
Writing -- review \& editing & L & S & S \\
\bottomrule
\end{tabular}
\caption{CRediT roles of the equal-contribution authors.}
\label{tab:credit}
\end{table}

\end{document}